\documentclass[manuscript,screen]{acmart}

\usepackage{lscape}
\usepackage{longtable}
\usepackage{xcolor}
\AtBeginDocument{%
  \providecommand\BibTeX{{%
    \normalfont B\kern-0.5em{\scshape i\kern-0.25em b}\kern-0.8em\TeX}}}

\setcopyright{acmcopyright}
\copyrightyear{2023}
\acmYear{2023}
\acmDOI{10.48550/arXiv.2210.06959}

\begin{document}

\title{A Survey on Explainable Anomaly Detection}

\author{Zhong Li}
\orcid{https://orcid.org/0000-0003-1124-5778}
\email{z.li@liacs.leidenuniv.nl}
\author{Yuxuan Zhu}
\orcid{https://orcid.org/0000-0001-5373-4452}
\email{y.zhu.12@umail.leidenuniv.nl}
\author{Matthijs van Leeuwen}
\orcid{https://orcid.org/0000-0002-0510-3549}
\email{m.van.leeuwen@liacs.leidenuniv.nl}
\affiliation{%
  \institution{Leiden Institute of Advanced Computer Science (LIACS), Leiden University}
  \streetaddress{Snellius Gebouw, Niels Bohrweg 1}
  \city{Leiden}
  \country{The Netherlands}
  \postcode{2333CA}}

\renewcommand{\shortauthors}{Li et al.}

\begin{abstract}
In the past two decades, most research on anomaly detection has focused on improving the accuracy of the detection, while largely ignoring the explainability of the corresponding methods and thus leaving the explanation of outcomes to practitioners. As anomaly detection algorithms are increasingly used in safety-critical domains, providing explanations for the high-stakes decisions made in those domains has become an ethical and regulatory requirement. Therefore, this work provides a comprehensive and structured survey on state-of-the-art explainable anomaly detection techniques. We propose a taxonomy based on the main aspects that characterize each explainable anomaly detection technique, aiming to help practitioners and researchers find the explainable anomaly detection method that best suits their needs.
  
\end{abstract}


%\ccsdesc[500]{Information systems~Information systems applications}


\maketitle

\section{Introduction}

An anomaly is an object that is notably different from the majority of the remaining objects. Depending on the specific application domain, an anomaly can also be called an outlier or a novelty. Moreover, it may also be known as an unusual, irregular, atypical, inconsistent, unexpected, rare, erroneous, faulty, fraudulent, malicious, unnatural, or strange object \citep{ruff2021unifying}. Except for a few works such as Reference \cite{ruff2021unifying}, the term \textit{outlier} is often used as a synonym for \textit{anomaly} in most research. For consistency, we will use the term \textit{anomaly} in this paper. 

Since the seminal work in \cite{knorr1998algorithms}, anomaly detection has been well studied and there exists a plethora of comprehensive surveys and reviews on it, including but not limited to References \cite{markou2003novelty,markou2003noveltyb,agyemang2006comprehensive,patcha2007overview,chandola2009anomaly,zimek2012survey,aggarwal2015outlier, chalapathy2019deep, boukerche2020outlier, pang2021deep}. 
In contrast, we only found a handful of surveys \citep{sejr2021explainable,panjei2022survey, yepmo2022anomaly} about the \emph{explainability} of anomaly detection methods. As suggested by {\color{black}Langone et al.} \cite{langone2020interpretable}, model explainability represents one of the main issues concerning the adoption of data-driven algorithms in industrial environments. More importantly, for applications in safety critical domains, providing explanations to stakeholders of AI systems has become an ethical and regulatory requirement \citep{voigt2017eu,european2020artificial}. However, after a thorough survey of academic publications on explainable anomaly detection, we found that existing surveys are either outdated, have missed some important work, or their proposed taxonomies are relatively coarse and therefore unable to characterize the increasingly rich set of explainable anomaly detection techniques available in the literature.

To address this gap in the literature, we conduct a comprehensive and structured survey on state-of-the-art explainable anomaly detection techniques and distill a refined taxonomy that caters to the increasingly rich set of techniques. Overall, this survey intends to provide both practitioners and researchers with an extensive overview of the different types of methods that have been proposed, with their pros and cons, and to help them find the explainable anomaly detection technique most suited to their needs. 

Note that some researchers \citep{montavon2018methods,broniatowski2021psychological,sipple2022general} distinguish between the terms `interpretation' and `explanation',  the terms `interpretable' and `explainable', and the terms `explainability' and `interpretability'. Specifically, {\color{black}Broniatowski} \cite{broniatowski2021psychological} defines explainability as \textit{a model's ability to provide a description of how its outcome came to be} and describes interpretability
as \textit{a human's ability to make sense from a given stimulus so that the human can make a decision}. Moreover, {\color{black}Sipple \& Youssef }\cite{sipple2022general} argue that explainability is \textit{ the algorithmic task of generating the explanation}, and interpretability is \textit{the cognitive task of merging the expert’s knowledge with the explanation to identify a unique diagnostic condition and to choose the appropriate treatment.} Considering that most researchers in data mining and machine learning treat explainability and  interpretability equally, we use those terms interchangeably throughout this paper. The next section will clarify what we mean exactly when we say that a technique is explainable.

\subsection{Methodology}
This survey aims to answer the following research questions and is structured accordingly:
\begin{itemize}
    \item[Q0] What is explainable anomaly detection and why should we care about it?
    \item[Q1] What are the most important aspects that characterize each explainable anomaly detection technique? On this basis, how to classify existing techniques? 
    \item[Q2] How do existing techniques interpret anomalies and what are the main differences between them?
    \item[Q3] What are the challenges and associated opportunities in explainable anomaly detection? 
\end{itemize}
In order to answer these research questions, we employ a comparative and iterative surveying procedure that consists of three cycles. In the first cycle, we employ a methodology consisting of two main phases:
\begin{itemize}
    \item Database Selection: 
    we select well-known scientific databases for literature collection, i.e., Google Scholar, IEEE Xplore, ACM Digital Library, DBLP, and Web of Science.
    \item Literature Selection: we select related research publications that were published between January 1998 to February 2022 using the following keywords: Interpretable/Interpret/Interpreting Anomaly Detection, Explainable/Explain/Explaining Anomaly Detection,  Interpretable/Interpret/Interpreting Outlier Detection, Explainable/Explain/Explaining Outlier Detection, Anomaly Interpretation, Anomaly Explanation, Outlier Interpretation,  Outlier Explanation. Other useful keywords are: Anomaly/Outlier Description, Anomaly/Outlier Characterization, Outlying Property Detection, Outlying Aspects Mining, Outlying Subspaces Detection.
\end{itemize}
In the second cycle, we inspect research publications that have been referenced by papers collected in the first cycle. In the third cycle, we exclude research publications that are irrelevant, not published in what we consider high-quality venues, or applications of existing methods to certain use cases. 

This survey is organised as follows. To answer Q0, Section 2 states the motivations for this work and the terminology used. Section 3 describes the proposed taxonomy for answering Q1.
Sections 4, 5, 6 and 7 survey existing techniques for explainable anomaly detection in a principled manner based on the proposed taxonomy, aiming to answer Q2. Section 8 discusses the open challenges and related opportunities of existing work, and then concludes this survey, answering Q3.

\section{The Need for Explainable Anomaly Detection}

This section introduces important terminology and concepts, such as anomalies and explainable anomaly detection, and explains why this is an important field of study.

\subsection{What Is An Anomaly?}

\begin{figure}
  \centering
  \includegraphics[width=15cm]{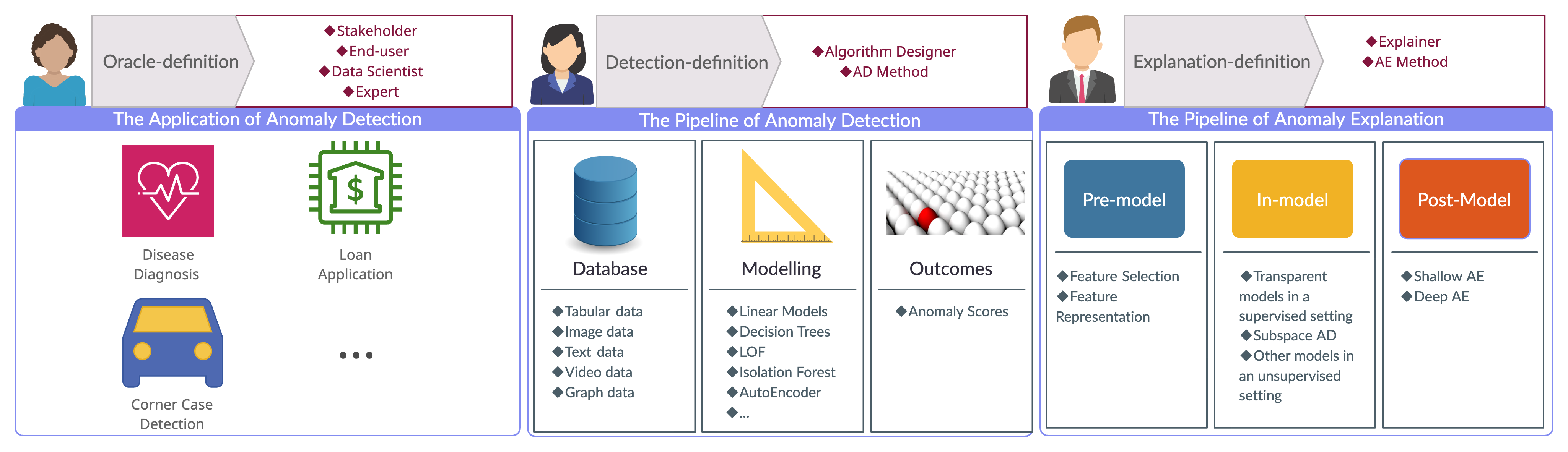}
  \caption{The procedure of anomaly analysis and the different roles involved in this procedure.}
  \label{Fig:ADprocedure}
\end{figure}

First of all, we need to define what an anomaly is. Inspired by {\color{black}Sejr \& Schneider-Kamp }\cite{sejr2021explainable}, we assume that there are three roles involved in an anomaly analysis task: 1) a/an
\textit{Stakeholder/End-user/Data Scientist/Expert} that uses the anomaly detection system; 2) an \textit{Algorithm Designer/Anomaly Detection Method} that does the actual anomaly detection; and 3) an \textit{Algorithm Explainer/Anomaly Explanation Method} that explains identified anomalies. These three roles are illustrated in Figure \ref{Fig:ADprocedure}. The different roles may have different definitions of what an anomaly is, and we distinguish those definitions as follows:

\begin{itemize}
    \item \textit{Oracle-Definition}: the `ideal' definition that defines the anomalies that the end-users of the anomaly detection system aim to detect. In other words, this definition defines the \textit{true anomalies} in the real-world application and thus strongly depends on the context and is often hard to formally/precisely formulate.
    \item \textit{Detection-Definition}: the anomalies that an anomaly detection model can actually capture. This definition is given explicitly or implicitly by the anomaly detection model or technique.
    \item \textit{Explanation-Definition}: describes why (and when) the anomaly explanation method considers an anomaly as anomalous.
\end{itemize}

For example, for a credit card fraud detection system, the end-users aim to detect fraudulent behaviour, which is defined as ``obtaining services/goods and/or money by unethical means'', including bankruptcy fraud, theft fraud, application fraud and behavioral fraud \citep{delamaire2009credit}. Therefore, the \textit{Oracle-Definition} is  ``behaviour that aims to obtain services/goods and/or money by unethical means''. However, a given credit card fraud system might only detect anomalous behaviours such as unprecedented high payments and/or payments at a never-before-seen location. Hence, the \textit{Detection-Definition} is ``unprecedented high payments and/or payments at a never-before-seen location'' and this is actually a theft fraud. Moreover, for an identified anomalous payment, the anomaly explanation method could generate the explanation ``the payment is flagged as anomalous because it happened at midnight'', which follows from the \textit{Explanation-Definition}. Clearly the \textit{Oracle-Definition}, the \textit{Detection-Definition}, and the \textit{Explanation-Definition} can be different from each other.

In general, the \textit{Oracle-Definition} is given based on domain knowledge, which is application-specific. From this point of view, there is no universal definition of an anomaly. A commonly accepted definition by {\color{black}Hawkins } \cite{hawkins1980identification} is that ``an outlier is an observation that deviates so much from other observations as to arouse suspicion that it was generated by a different mechanism''. As this is informal, each specific anomaly detection model has its own definition of an anomaly, either explicitly or implicitly. For example, KNN \citep{ramaswamy2000efficient} defines objects with `far' $k$-nearest neighbours as anomalies, LOF \citep{breunig2000lof} treats objects with a low local density as anomalies, and Isolation Forest \citep{liu2008isolation} considers `easily isolated' objects as anomalies. Importantly, this \textit{Detection-Definition} definition can be different from the  \textit{Oracle-Definition}, which may lead to problems. For example, an anomaly detector may miss relevant anomalies while detecting `anomalies' that are uninteresting to end-users. Moreover, depending on the technique used to explain an anomaly, the \textit{Detection-Definition} and \textit{Explanation-Definition} can also be different, especially when the explanation approach does not reflect the decision-making process behind the anomaly detection model.

\subsection{What is Explainable Anomaly Detection?}

According to {\color{black}Doshi-Velez \& Kim }\cite{doshi2017towards}, interpretability or explainability is defined as the ability to explain or provide meaning to humans in understandable terms. Moreover, {\color{black}Arrieta et al. }\citep{arrieta2020explainable} define Explainable Artificial Intelligence (XAI) as ``Given an audience, an explainable Artificial Intelligence is one that produces details or reasons to make its functioning clear or easy to understand.'' Further, {\color{black}Murdoch et al. }\cite{murdoch2019definitions} define interpretable or eXplainable Machine Learning (XML) as ``the extraction of relevant knowledge from a machine learning model concerning relationships either contained in data or learned by the model'', where the knowledge is considered relevant if it provides insight into the problem faced by the target audience. Accordingly, we define eXplainable Anomaly Detection (XAD) as \textit{the extraction of relevant knowledge from an anomaly detection model concerning relationships either contained in data or learned by the model}, where the knowledge is considered relevant if it can provide insight into the anomaly detection problem investigated by the end-user. Hereinafter, we utilize XAI and XML interchangeably as they practically mean the same within the scope of this manuscript.

%Reference \cite{miller2019explanation} defines XAI as a human-agent interaction problem, which is the intersection of Artificial Intelligence, Human-Computer Interaction (HCI), and Social Science (including Philosophy, Cognitive Science, and Social Psychology). As a subfield of XAI, XAD can also be regarded as the intersection of those three domains. Therefore, in addition to considering the computational problems in XAD, we should also consider problems such as \textit{how do humans understand an explanation}, \textit{what kind of explanations are human-understandable}, and \textit{how do humans interact with machines to understand explanations?} As pointed out in \cite{miller2019explanation}, for more than two decades, cognitive science has studied how humans generate explanations and assess their quality. For instance, they found that people can better understand an explanation if the explanation is presented in a contrastive manner with counterfactual event. Overall, it is important to investigate the cognitive aspects of explainability as well as  human-in-the-loop design of explainability. However, due to space constraints, we consider the HCI \citep{smits2022panda} and social science \citep{miller2019explanation,miller2021contrastive} aspects of XAD to be beyond the scope of this survey. Instead, we will mainly review the artificial intelligence aspect (namely algorithmic design) of XAD. 

{\color{black}Miller } \cite{miller2019explanation} defined XAI as a human-agent interaction problem at the intersection of Artificial Intelligence, Human-Computer Interaction (HCI), and the Social Sciences (including Philosophy, Cognitive Science, and Social Psychology). Being a subfield of XAI, XAD can also be situated at the intersection of those three domains. Therefore, in addition to considering different XAD tasks and problems together with their algorithmic and computational challenges, it would also be of interest to consider questions such as \textit{how do humans understand an explanation}, \textit{what kind of explanations are human-understandable}, and \textit{how do humans interact with machines to understand explanations?} Thoroughly addressing these questions, however, would require substantial additional coverage and analysis of the literature; to maintain a clear scope and prevent the survey from becoming even longer, we will not address these questions. Instead, we refer to recent papers for perspectives from HCI \citep{smits2022panda} and social science \citep{miller2019explanation,miller2021contrastive}, and leave a broader discussion of these aspects to a future article.

The anomaly analysis process consists of two equally important tasks, namely \emph{anomaly detection} and \emph{anomaly explanation}. Anomaly explanation refers to the process of finding out why an anomaly is considered anomalous. Because the terms \textit{anomaly} and \textit{outlier} are used interchangeably, anomaly explanation is also known as outlier explanation, outlier interpretation, outlier description, outlier characterization, outlying property detection, outlying aspects mining, outlying subspaces detection, object explanation, and promotion analysis. 

An anomaly can be identified by an anomaly detection algorithm or otherwise become known (e.g., from an expert). 

\begin{itemize}
    \item \textbf{Case 1 (Model)} If an anomaly is identified by an anomaly detection algorithm, \textit{XAD} aims to explain the anomaly by making the anomaly detection method interpretable. Specifically, there exist many approaches to make an anomaly detector interpretable. If the anomaly detector is intrinsically interpretable (e.g., logistic regression, shallow decision trees, rule-based models, etc.), it is relatively easy to deduce why the anomaly is flagged as anomalous. In contrast, if the anomaly detector is not intrinsically interpretable (e.g., Isolation Forest \citep{liu2008isolation}, RNN \citep{salehinejad2017recent}, CNN \citep{gorokhov2017convolutional}), post-hoc XAI techniques such as SHAP \citep{lundberg2017unified}, LIME \citep{ribeiro2016should}, and Anchors \citep{ribeiro2018anchors} can be use to interpret the anomaly detector, namely to describe why it makes certain decisions. In this case, we aim to make the  \textit{Detection-Definition} and \textit{Explanation-Definition} consistent.
    
    \item \textbf{Case 2 (Data)} If an anomaly is identified by an expert, an anomaly explanation method can only aim at explaining why the given data instance is anomalous, extracting no knowledge from any anomaly detection models. In this case, we attempt to make the \textit{Oracle-Definition} (if any) and \textit{Explanation-Definition} consistent. However, it is also possible that the expert obtains the anomaly by running an existing anomaly detection algorithm, but the design of the algorithm is unavailable to the expert for some reasons (such as confidentiality). Hence, the \textit{Explanation-Definition} may be different from the \textit{Detection-Definition} (which is not known). %Such an anomaly explanation method, capable of detecting anomalies and providing explanations at the same time, can be considered as a surrogate model for the unavailable anomaly detection model. 
\end{itemize}

In short, the biggest difference between these two cases is about what to explain: the model (and possibly the data) or just the data. \textbf{Case 1} is centered around anomaly detection models. If we can understand how the anomaly detection model makes decisions, as a by-product, we can easily explain why an anomaly is flagged as anomalous by the model. In contrast, \textbf{Case 2} focuses on anomalies and aims at explaining why they are anomalous where the detection model is not available. The anomaly explanation methods corresponding to this case can be considered as surrogate methods for the unavailable anomaly detection models. For completeness, we will consider both cases in this survey.

\subsection{Why Should We Care About XAD?}

Due to the widespread application of anomaly detection in many domains, the interpretability of corresponding methods has become increasingly important \citep{panjei2022survey}. For example, anomaly detection algorithms are being used to diagnose diseases in healthcare \citep{ukil2016iot}. In financial services, many banks use anomaly detection methods to detect abnormal behaviour in credit card transactions \citep{ahmed2016survey}. In addition, the self-driving car manufacturing industry applies anomaly detection algorithms on camera data to detect corner cases \citep{bogdoll2022anomaly}. In other safety-critical areas---such as spacecraft design---anomaly detection algorithms are used to detect sensor faults \citep{fuertes2016improving}. As we can see, anomaly detection systems for high-stakes decisions are deeply impacting our daily lives and society. One natural question is, \textit{how can we trust these systems without understanding and validating the underlying rationale of the involved anomaly detection components?} For this reason, XAD aims to not only provide accurate anomaly detection results, but also to provide tangible explanations of why a specific object is detected as an anomaly \citep{pang2021toward}. 

Providing anomaly detection results with corresponding explanations can help gain the trust of end-users in anomaly detection systems. Moreover, the explanations can also assist end-users to validate the anomaly detection results in unsupervised settings. Even more, explanations can potentially enable end-users to find the root causes of anomalies and thereby take remedial or preventive actions.

For a long time, however, the anomaly detection community has mainly focused on detection accuracy, largely ignoring the interpretation of corresponding decisions. {\color{black}For instance, Micenková et al. }\cite{micenkova2013explaining} criticise that ``almost all existing algorithms stop at the point of providing anomaly ranking and leave the user without any explanation of why some data points deviate and how.'' Additionally, {\color{black}Dang et al. }\cite{dang2013local} indicate that ``although there is a large number of techniques for discovering global and local anomalous patterns, most attempts focus solely on the aspect of outlier identification, ignoring the equally important problem of outlier interpretation.'' {\color{black}Aggarwal }\cite{aggarwal2015outlier} also points out that ``only few outlier detection studies considered providing some qualitative information to explain the form of outlierness.'' {\color{black}Similarly, Vinh et al. } \cite{vinh2016discovering} argue that ``current outlier detection techniques do not usually offer an explanation as to why the outliers are considered as such, or in other words, pointing out their outlying aspects.''

In summary, the anomaly detection community has long been paying more attention to \textit{giving correct answers} rather than \textit{providing explanations} or---even better---\textit{providing correct explanations}. With more and more applications or potential applications of anomaly detection in high-risk decision-making systems, it has become crucial to gain or increase humans' trust in and acceptance of anomaly detection techniques. For this it is important to provide correct answers with correct explanations, i.e., to avoid the \textit{Clever Hans Phenomenon} \citep{lapuschkin2019unmasking} that---in this context---refers to anomaly detection models utilising spurious correlations and patterns in the data to identify anomalies. Although the identified anomalies are true, these correlations or patterns may be incorrect or undesirable (e.g., violating the laws of physics). Such provably incorrect explanations are unacceptable to end-users and would only harm trust.

\subsection{What is A Good XAD Method?}
%We have indicated the importance of generating explanations for anomaly detection models or given anomalies. Beyond that, another question arises, which is \textit{how should we trust your explanations without quantifying their quality?} Therefore, it is crucial to evaluate the quality of generated explanations.

Once explanations are generated by an XAD method, how can one trust them? A natural first step is to evaluate the quality of generated explanations. {\color{black}Studies relevant to this have been conducted in the realm of XAI. For instance, references }\cite{guidotti2018survey,belle2021principles} analyze the XAI literature and propose important properties that should be considered when designing an XAI technique. Next, {\color{black}Barbado et al. } \cite{barbado2022rule} defines some criteria to evaluate rule-extraction-based explanation techniques. {\color{black}Moreover, Zhou et al. } \cite{zhou2021evaluating} performs a survey on the quality evaluation of machine learning explanations. Recently, {\color{black}Sipple \& Youssef }\cite{sipple2022general} proposes four desiderata for anomaly explanation methods as well as a method for comparing different explanations. However, there is no consensus on what a good XAD technique should be. Based on related work on XAI, we find the following properties to be especially relevant when designing or choosing an XAD technique: 
\begin{itemize}
    \item Accuracy: how accurate is the prediction of unseen anomalous instances as anomalies;
    \item Fidelity: consistency of \textit{Oracle-Definition},  \textit{Detection-Definition}, and \textit{Explanation-Definition};
    \item Comprehensibility: to what extent are the explanations understandable to the end-users;
    \item Generality: does the technique have special requirements for data type, data size, anomaly detection model type, anomaly detection model size, training regimes or training restrictions;
    \item Scalability: does it scale to large input data size and/or a large model;
    \item Complexity: how many hyper-parameters need to be set by end-users.
\end{itemize}

The practical implementation and evaluation of XAD techniques is largely dependent on the application domain and end-users, and is therefore out of the scope of this survey.

\section{A Taxonomy of Explainable Anomaly Detection Methods} \label{sec:Taxonomy}

\begin{figure}
\centering
\includegraphics[width=15.2cm]{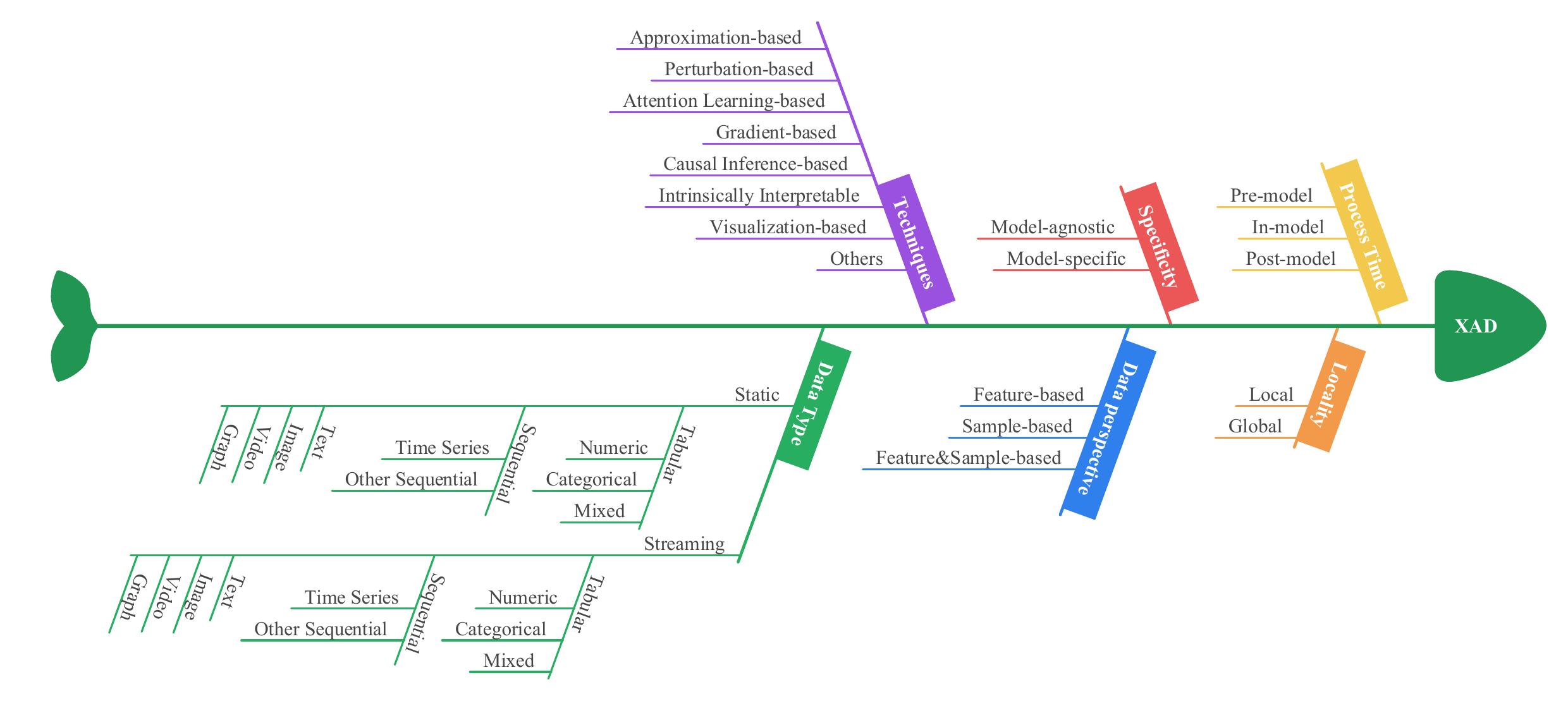}
\caption{XAD taxonomy based on six criteria (colored text boxes). Most existing XAD methods fall into one category for each of the six criteria.}
\label{Fig:Tree}
\end{figure}

Before we introduce the taxonomy that we propose for the field of explainable anomaly detection (XAD), we first briefly review existing surveys and taxonomies.

\subsection{Related Work}

Compared to the abundance of taxonomies of anomaly detection methods, including but not limited to these surveys throughout the years \citep{hodge2004survey,chandola2009anomaly,gupta2013outlier,agrawal2015survey,chalapathy2019deep,pang2020deep}, the categorization of anomaly explanation methods involving XAD techniques has received relatively little attention so far \citep{vinh2016discovering,samariya2020comprehensive,panjei2022survey,yepmo2022anomaly,sejr2021explainable}.

{\color{black}We discuss the four most notable existing categorizations. Vinh et al. }\cite{vinh2016discovering} for the first time subdivided anomaly explanation approaches into two categories: \textit{Feature selection based approaches} that transform the anomaly explanation task into the classical problem of feature selection for classification, and \textit{Score-and-search approaches} that compare the outlyingness degree of an anomaly across all subspaces followed by inspecting the subspace with the highest anomaly score. To the best of our knowledge, {\color{black}Samariya et al. }\cite{samariya2020comprehensive} was the first work dedicated to the survey of anomaly explanation methods. They also subdivided related techniques into three categories: \textit{Score-and-Search based approaches}, \textit{Feature selection based approaches}, and \textit{Hybrid approaches}. More recently, {\color{black}Panjei et al. }\cite{panjei2022survey} introduced a survey on anomaly explanation, wherein they divided relevant techniques into three categories: \textit{Importance Levels of Outliers}, \textit{Causal Interactions Among Outliers}, and \textit{Outlying Attributes}. {\color{black}Meanwhile, Yepmo et al. } \cite{yepmo2022anomaly} also presented a review of anomaly explanation methods, categorizing existing techniques into four groups, namely \textit{Explanations by Feature Importance}, \textit{Explanations by Feature Values}, \textit{Explanations by Data Points Comparison}, and \textit{Explanations by Structure Analysis}. Finally, Reference \cite{sejr2021explainable} is also closely related, wherein they have discussed what anomaly explanations are, who needs those explanations, and why there are different types of anomaly explanations.

After a thorough survey of the scientific literature on XAD techniques, we find that existing surveys are less comprehensive than we aim to be in this manuscript. Specifically, each of the above surveys contains no more than 40 relevant works in the field. In contrast, our survey has investigated more than 150 relevant papers. In addition, we find the existing taxonomies to be relatively coarse and sometimes not intuitive. For example, although anomaly score is a very natural ranking of outlying degree, {\color{black}Panjei et al. }\cite{panjei2022survey} particularly treat anomaly ranking as a subcategory of anomaly explanation methods. Further, although \textit{Explanations by Feature Importance} and \textit{Explanations by Feature Values} mainly differ in the granularity of provided explanations, {\color{black}Yepmo et al. }\cite{yepmo2022anomaly} regard them as two distinct categories. In brief, existing surveys only partially cover existing research, and the proposed taxonomies are insufficient to characterize the increasingly rich field of XAD. For this reason, we perform a comprehensive and structured survey on state-of-the-art XAD techniques. As new articles are published at a rapid pace, we do not claim to have covered all relevant research publications. Furthermore, as we intend to include a wide spectrum of XAD methods, we cannot describe each method in detail. Meanwhile, a refined taxonomy, distilled from existing surveys on XAI techniques, is presented below and used to categorize XAD methods.

\subsection{Proposed Taxonomy}

Similar to how anomaly detection is an important part of machine learning and data mining, we argue that XAD is also an important constituent of what is nowadays called XAI. XAI has received extensive attention in the past few years due to the emergence and prevalence of black-box models such as deep neural networks. After carefully scrutinizing existing surveys on XAI \citep{gilpin2018explaining,dovsilovic2018explainable,carvalho2019machine,arrieta2020explainable,linardatos2020explainable,belle2021principles,burkart2021survey}, we found that some criteria are often used to categorize existing XAI techniques. Capitalizing on these findings, we propose six main criteria to taxonomize existing XAD techniques. 

First of all, according to the anomaly detection pipeline as shown in Figure \ref{Fig:ADprocedure}, we can subdivide XAD techniques into three categories, namely \textit{Pre-model techniques}, \textit{In-model techniques} and \textit{Post-model techniques}. Specifically, \textit{pre-model techniques}, also known as \textit{ante-hoc techniques}, are constructed and implemented before the anomaly detection process. Techniques such as filter feature selection methods belong to this category. \textit{In-model techniques} use inherently interpretable models and can therefore provide explanations without additional or with little efforts when performing anomaly detection. For example, anomaly detection methods based on linear regression, which can simultaneously report the coefficients of the corresponding features, fall into this category. In contrast, \textit{post-model techniques}, also known as \textit{post-hoc techniques}, attempt to explain the decisions made by an anomaly detection model after the construction and implementation of the detection model or when anomalies are obtained from an oracle. For instance, SHAP-based interpretation methods \citep{lundberg2017unified} are part of this category. 

Second, we distinguish XAD techniques based on whether they provide a \textit{global explanation} or \textit{local explanation}. Specifically, a \textit{global explanation} is based on the understanding of the complete `model logic' or some important properties of the anomaly detection model, being able to explain how all decisions are made. In contrast, a \textit{local explanation} explains why a specific object is anomalous or how a specific decision is made.

Third, XAI techniques can be further subdivided into \textit{model-agnostic} approaches that can be applied to any anomaly detection model, and \textit{model-specific} approaches that are only applicable to specific anomaly detection models.

Fourth, two aspects of a tabular dataset can be used to generate explanations, i.e., a tabular dataset has features and samples. Therefore, we can subdivide techniques into three subcategories:
\begin{itemize}
    \item \textit{Feature-based methods} provide explanations based on features. This group of methods generally indicates which features are important and/or the corresponding values of investigated anomalies. Specifically, \textit{subspace} (e.g., a subset or unordered features), \textit{a set of subspaces} (e.g., a set of feature pairs), \textit{feature importance} (e.g., assigning a score or an order to each feature), and \textit{feature values} (e.g., rare combination of feature values) fall into this subcategory. Particularly, some studies attempt to define a set of rules based on a subset of features and their corresponding values, resulting in so-called patterns. Meanwhile, for sequential data such as time series,  a pattern consisting of a collection of consecutive observations is usually leveraged to detect and explain anomalies. Each observation can be regarded as a feature or a sample depending on the context. For simplicity, we call them \textit{pattern-based methods}, 
    but they are still essentially \textit{feature-based methods}.
    \item \textit{Sample-based methods} generate explanations based on samples. This type of method typically compares the abnormal object directly to normal objects to demonstrate differences. For instance, \textit{local neighbourhood} (e.g., the nearest objects, which may be normal or abnormal, to an anomaly), \textit{counterexample} (e.g., the nearest normal object to an anomaly), and \textit{contextual anomalies} (e.g., the nearest cluster to an anomaly) belong to this subcategory. Moreover, \textit{exception analysis} in Reference \cite{guidotti2018survey} and  \textit{representative examples} in References \cite{belle2021principles,tan2020tree}
    also fall under this category.
    \item \textit{Feature and Sample-based methods} leverage both aspects.
\end{itemize}

Fifth, based on the specific techniques used to generate explanations, we can categorize models into the following subcategories, which are not mutually exclusive: 
\begin{itemize}
    \item \textit{Approximation-based methods}, which approximate or mimic complex models with simpler ones that are much easier to interpret. They are also called surrogate models. Examples include LIME \citep{ribeiro2016should} and Anchors \citep{ribeiro2018anchors}.
    \item \textit{Perturbation-based methods}, which examine the influence of output via input changes to generate explanations. Examples include Anchors \citep{ribeiro2018anchors}.
    \item \textit{Reconstruction Error-based methods}, which use reconstruction errors to explain anomalies. Examples include SHAP-based methods \cite{antwarg2019explaining}.
    \item \textit{Attention Learning-based methods}, which use attention learning to localise anomalies.  Examples include Reference \cite{brown2018recurrent}  and Reference \cite{venkataramanan2020attention}.
    \item \textit{Gradient-based methods}, which measure feature contribution on midput (intermediate outputs) or outputs through back-propagation. Examples include Layer-wise Relevance Propagation  \cite{kauffmann2020towards,sipple2020interpretable,pang2021explainable}. Note that some of these methods may also be Reconstruction Error based. 
    %\item \textit{Entropy-based methods}, which perform model inference with information theory.
    \item \textit{Causal Inference-based methods}, which analyze the causal relations between objects and/or features to explain anomalies. Examples include Reference \cite{liu2011discovering}. 
    \item \textit{Visualization-based methods}, which use plots to explain anomalies. Examples include Reference \cite{liznerski2020explainable},  which uses heatmaps that is a kind of saliency masks. Note that many other techniques also leverage visualization to explain anomalies.
    \item \textit{Intrinsically Explainable methods}. The above mentioned subcategories are mainly post-model techniques that are used to explain deep learning based anomaly detection models. Meanwhile, there are in-model techniques that make the anomaly detection model intrinsically explainable. Examples include Rule-based models \cite{he2005fp}.
    \item \textit{Miscellaneous other methods}. We assign other techniques into this subcategory by indicating their specific technique. Examples include  Pattern Compression \cite{smets2011odd}, and Subspace Anomaly Detection \cite{savkli2021random}. 

\end{itemize}

Sixth and last, we also indicate the types of data to which each XAD technique can be applied. Specifically, the data type can be static or streaming. Furthermore, it can be tabular (numeric, categorical, or mixed), {\color{black}sequential (time series, other sequential)}, image, text, video, or graph data. 

Our overall proposed taxonomy is presented in Figure \ref{Fig:Tree}: each of the six criteria can be used to categorize an XAD method. Together these six `dimensions' can be used to provide a detailed characterization of an existing XAD method, or---the other way around---to find XAD methods satisfying certain requirements.

\subsection{Organisation of the Literature Review}

As described in the previous subsection and shown in Figure \ref{Fig:Tree}, our taxonomy employs six criteria. To organize our survey by these six criteria, however, we would have to introduce many section levels and some subsections would be much longer than others. We will therefore use another structure for the literature review in the following sections, which we will explain next. We will still make ample use of our proposed taxonomy: to partially structure the individual sections, to characterize the methods that we describe, and to provide full characterization of all methods in a large overview table at the end of each section.

\begin{figure}
\centering
\includegraphics[width=15cm]{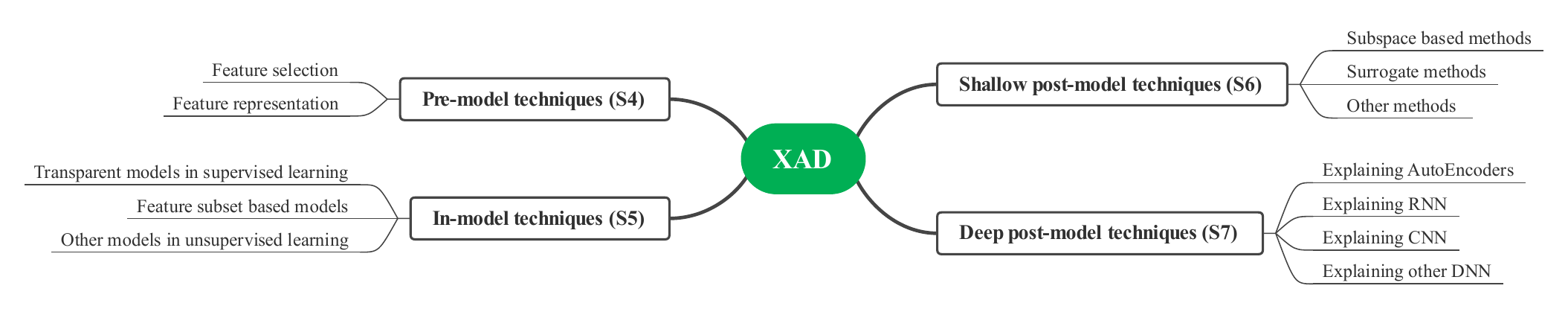}
\caption{Structure of the core of this survey, i.e., Sections 4--7 (indicated by S4--7).}
\label{Fig:Structure}
\end{figure}

We use the first main criterion to classify \textit{pre-model techniques} (S4), \textit{in-model techniques} (S5), and \textit{post-model techniques} (S6-7) into different sections. As there are so many post-model techniques, we split those into deep learning based methods (S7) and other, `shallow' methods (S6). Next, we use the characteristics of each of these categories to define subsections. That is, the \textit{pre-model techniques} section consists of subsections for \textit{feature selection} and \textit{feature representation}. Meanwhile, the \textit{in-model techniques} section includes subsections for \textit{transparent models in supervised learning}, \textit{feature subset based models}, and \textit{other models in unsupervised learning}. The \textit{shallow post-model techniques} section has subsections for \textit{subspace based methods}, \textit{surrogate methods}, and \textit{miscellaneous methods}. Finally, the \textit{deep post-model techniques} section contains subsections on \textit{explaining AutoEncoders}, \textit{explaining RNNs}, \textit{explaining CNNs}, and \textit{explaining other DNNs}.

\section{Literature Review on pre-model techniques}
Opaque models are often criticized for their inexplicability. However, the features used as inputs to models are as critical as, if not more than, the type of models in producing explainable results. In other words, by having more meaningful and informative features whilst retaining fewer irrelevant features, we can build simpler models to learn the relationships exhibited in the data while ensuring comparable anomaly detection accuracy.

Therefore, this section reviews papers that leverage XAD techniques \emph{before} the anomaly detection process. Specifically, the following pre-model techniques are investigated:
\begin{itemize}
    \item Feature selection methods that select a subset of original features for anomaly detection;
    \item Feature representation methods that learn a set of high-level and human-understandable feature representations for anomaly detection.  
\end{itemize}

\subsection{Feature Selection For Anomaly Detection}
\label{Sec:PreModel1}

%[Todo] Read several papers on feature selection for AD

{\color{black}Siddiqui et al. }\cite{siddiqui2019sequential} point out that the effort required to investigate an anomaly is usually proportional to the number of features that describe it. Therefore, dimensionality reduction techniques---including feature projection and feature selection methods---can be applied to reduce the number of features that describe an object, thereby facilitating anomaly explanation. However, feature projection methods such as Principal Component Analysis convert the original features into a new set of features, sacrificing interpretability. In contrast, feature selection methods retain a subset of original features that contain the most important information, greatly improving the interpretability and effectively alleviating the \textit{curse of dimensionality} problem in high-dimensional data.

There exist very limited unsupervised feature selection methods for anomaly detection. Specifically, {\color{black}Pang et al. }\cite{pang2016outlier} and {\color{black}Pang et al. }\cite{pang2016unsupervised} propose two filter-based unsupervised selection methods, namely CBRW\_FS and CBRW, which select a subset of features independently from subsequent anomaly detection methods. These two methods work only on categorical data through modeling the feature-value couplings. {\color{black}By assuming strong similarities between rare instances, He \& Carbonell }\cite{he2010co} design an optimization framework to jointly select features and instances for anomaly detection on categorical data. However, this assumption is usually not satisfied since  anomalies are often isolated and thus distinct from each other. 

{\color{black}Meanwhile, Noto et al. }\cite{noto2012frac} and {\color{black}Paulheim \& Meusel }\cite{paulheim2015decomposition} try to find a relevant feature subset for anomaly detection by exploring the correlations between features. They assume that anomalies are those instances that violate the normal dependencies between features. Therefore, only features that are related to other features are considered relevant for anomaly detection. Unfortunately, this anomaly definition is not applicable to many benchmark anomaly detectors. Moreover, Isolation Forest \citep{liu2008isolation} can also be used to select a subset of features for anomaly detection. The isolation forest based feature selection method, IBFS \citep{yang2019isolation}, simply selects features that contribute the most to the outlyingness of anomalies reported by the Isolation Forest method. To our knowledge, this is the first unsupervised feature selection method specifically designed for \textit{generic} anomaly detection in numeric data. The above three methods are all filter-based, which independently select subsets of features regardless of subsequent anomaly detection methods. Consequently, suboptimal or completely irrelevant features may be selected for anomaly detectors.

 A platform information technology (PIT) system is a system capable of connecting and communicating with other systems, subsystems and devices. To detect attacks in PIT systems, {\color{black}Morris }\cite{morris2019explainable} proposes to use Principal Component Analysis (PCA) or Independent Component Analysis (ICA) to reduce the number of features considered, thereby promoting interpretability in the subsequent anomaly detection process. Moreover, he suggests using ensemble learning based methods such as Random Forests to detection anomalies after the dimensionality reduction process. However, every feature obtained using PCA or ICA is a combination of the original features and is therefore no longer interpretable.

 Some feature selection methods are interleaved with the anomaly detection process, rather than being applied before the anomaly detection process. We call such methods wrapper or embedded feature selection methods depending on their implementations, and  will introduce them in the next section.

\subsection{Feature Representation For Anomaly Detection}

Due to the complexity entailed in data such as time series, image, video, etc., deep neural network (DNN) based methods have shown superiority in detecting anomalies in these data. However, DNN-based models are notoriously known for their complexity, which implies uninterpretability. {\color{black}To alleviate this problem, Chen et al. }\cite{chen2018scene} and {\color{black}Wu et al. }\cite{wu2021explainable} indicate that using high-level and human-understandable feature representations for anomaly detection can reduce the complexity of subsequent anomaly detection models, thereby improving their interpretability.

{\color{black}Examples can be observed in the domain of time series anomaly analysis. For instance, Ramirez et al. }\cite{ramirez2019computational} introduce an interpretable anomaly detection and classification framework to analyze human gait. Specifically, they first harness symbolic representations such as Piecewise Aggregate Approximation to represent the collected multivariate time series data. Particularly, they consider the symbolic abstraction of the data as the core of their XAD framework, enhancing interpretability of the results via feature reduction. Second, they apply two discords based anomaly detection methods, viz. HOT-SAX \citep{keogh2005hot} and RRA \citep{senin2015time}, to discovery anomalies, respectively. Third, they determine the final anomalies based on the consensus of these two detection algorithms.

{\color{black}Instead of using symbolic representations, Dissanayake et al. }\cite{dissanayake2020robust} investigate the importance of heart sound segmentation {\color{black}and feature extraction} for detecting abnormal heart sound. They suggest that an automated detection method usually consists of three steps: Segmentation, Feature Extraction, and Classification. First, they apply the model proposed by {\color{black}Fernando et al. }\cite{fernando2019heart} to perform segmentation. Particularly, the segmentation is based on a feature representation called Mel-Frequency Cepstral Coefficients (MFCCs). They argue that pre-extracted feature representations such as MFCCs or spectrogram are commonly used in medical domain as they are closely related to the original signal. One can gain important insights into the model prediction results if explaining the feature representations in conjunction with the signal. Second, they utilise a Convolution Neural Network (CNN) encoder to extract features. Third, they construct a Multilayer Perceptron Network (MLP) model to perform anomaly detection. Moreover, to interpret an anomaly, they combine Shapley values and Occlusion maps \citep{zeiler2014visualizing} to investigate how input features impact the prediction.

{\color{black}Schlegl et al. }\cite{schlegl2021scalable} construct a deep neural network-based model that can learn interpretable feature representations from unlabeled time series, facilitating the evaluation and deployment of subsequent anomaly detection algorithms. First, they set up a so-called \textit{deviation convolution} based model to learn  characteristic shapes of normal time series, wherein they impose a separating constraint on the neural network to make it interpretable. Second, they feed these human-interpretable shapes to a convolutional-RNN AutoEncoder, which attempts to reconstruct the input shapes while minimising the reconstruction errors. Therefore, a test instance with a large reconstruction error is considered anomalous.

{\color{black}In the field of video anomaly analysis, Wu et al. }\cite{wu2021explainable} propose a Denoising AutoEncoder (DAE) based model combined with SHAP to detect and explain anomalies in videos. Since uninterpretable feature representations hide the decision-making process, they first leverage pretrained Convolutional Neural Network (CNN) models to extract high-level concept and contextual features. Second, they train a DAE model based on these features to predict the video frame. On this basis, a test instance is considered anomalous if its actual frame is significantly different from its predicted frame. Third, they apply kernel SHAP \citep{lundberg2017unified} to find input features which cause the anomaly.

%To add more work:
%1). Similarly, Reference \cite{rad2021explainable} introduces HIDR that utilises an AE model to detect anomalies and then leverages LIME to help explain anomalies.
%\cite{rad2021explainable} & S (AEs) & F & Gradient based (EXstream) + Approximate (LIME) + Others (MacroBas) & Streaming MTS  & L & Comparisons of three XAD methods & ------\\
%2) Unsupervised Feature Selection for Outlier Detection on Streaming Data to Enhance Network Security
%3) A filter-based feature selection model for anomaly-based intrusion detection systems
%4) Spectral ranking and unsupervised feature selection for point, collective, and contextual anomaly detection
%5) A hybrid feature selection scheme and self-organizing map model for machine health assessment
%6) Outlier Detection Ensemble with Embedded Feature Selection
%7) Unsupervised feature selection and cluster center initialization based arbitrary shaped clusters for intrusion detection
%8) Unsupervised feature selection for outlier detection in categorical data using mutual information
%9) Simultaneous feature selection and outlier detection with optimality guarantees
%10) Information-value-based feature selection algorithm for anomaly detection over data streams
%11) Feature selection for anomaly–based network intrusion detection using cluster validity indices
%12) Unsupervised Feature Selection Method for Intrusion Detection System
%13) GPU-Accelerated Feature Selection for Outlier Detection Using the Local Kernel Density Ratio

\subsection{Summary}
As shown in Table \ref{tab:summary_pre}, all pre-model XAD techniques are model-agnostic except for Reference \cite{ramirez2019computational}. In other words, most pre-model XAD techniques can be applied to any subsequent anomaly detection methods. However, fully decoupling the feature selection or feature representation learning from the subsequent anomaly detection methods may lead to sub-optimal detection accuracy.

Furthermore, most reviewed pre-model XAD techniques are feature-based with the exception that {\color{black}He \& Carbonell }\cite{he2010co} also perform instance selection to improve interpretability. Importantly, all pre-model XAD techniques can provide global explanations in the sense that they render the subsequent anomaly detection models more transparent and interpretable by preserving less irrelevant or redundant features, or providing  human-understandable feature representations.

The ultimate goal of using XAD techniques is to ensure that the entire pipeline of anomaly detection is human-understandable. However, we note that high-level and human-understandable feature representations are usually obtained by an opaque model, such as a pre-trained CNN model in Reference \cite{dissanayake2020robust}, which somewhat offsets the benefits of using interpretable feature representations for anomaly detection. 

Moreover, it can be seen that the reviewed feature selection and feature representation techniques are model-based feature engineering methods, which only leverage machine learning techniques. However, one can employ domain-knowledge based feature engineering methods to extract features. For instance,  {\color{black}Murdoch et al. }\cite{murdoch2019definitions} point out that combining exploratory data analysis tools with domain knowledge is helpful for extracting meaningful features, thereby improving the interpretability of subsequent anomaly detection.

\begin{table}
\caption {Summary of pre-model XAD techniques. \textit{Spec} indicates whether a method is model-agnostic (A) or model-specific (S). \textit{Pers} specifies whether a method is feature-based (F), sample-based (S) or pattern-based (P). \textit{Tech} indicates the techniques used in each method. \textit{Data} indicates the data type for which the method is applicable (TN: Tabular Numeric; TC: Tabular Categorical; TM: Tabular Mixed; UTS: Univariate Time Series; MTS: Multivariate Time Series; ES: Event Sequence). \textit{Loc} shows whether a method provides a local explanation (L) or global explanation (G). \textit{Pros} and \textit{Cons} describe advantages and disadvantages of each method, respectively.}
\label{tab:summary_pre}
\small\begin{tabular}{p{0.03\textwidth}p{0.02\textwidth}p{0.05\textwidth}p{0.23\textwidth}p{0.08\textwidth}p{0.05\textwidth}p{0.18\textwidth}p{0.18\textwidth}
}
\hline Ref & Spec & Pers & Tech & Data &Loc  & Pros & Cons\\
\hline
\cite{pang2016outlier} & A & F  & Feature selection & Static TC & G  & Handles noisy features well  & Only applicable to categorical data \\
\cite{pang2016unsupervised} & A & F & Feature selection & Static TC & G & Linear time complexity to data size & Only applicable to categorical data \\
\cite{he2010co} & A & F \& S & Feature selection + Instance selection & Static TC & G & Jointly selects features and instances for AD & Assumes strong similarities between rare instances\\
\cite{noto2012frac} & A & F & Feature selection & Static TN & G  & Robust to noisy and high-dimensional data & Only explores correlations between features\\
\cite{paulheim2015decomposition} & A & F  &  Feature selection & Static TN & G  & Changes unsupervised AD into supervised AD  & Only explores correlations between features\\
\cite{yang2019isolation} & A & F & Feature selection  & Static TN &  G & Applicable to generic AD for numeric data  & Selects features without considering subsequent AD methods\\
\cite{morris2019explainable} & A & F & Feature selection & Static TN & G  & Applicable to generic AD  & Obtained features are not interpretable \\
\cite{wu2021explainable} & A  & F & Pretrained CNN models to extract high-level concept and contextual features; VAE + SHAP & Static video & L \& G   & Extracted features are easy to understand & Weak interpretability due to the opacity of CNN \\
\cite{ramirez2019computational} & S  & P & Symbolic representation using PAA & Static MTS &  G & Enables human-in-the-loop & Only applicable to symbolic based AD such as HOT-SAX and RRA\\
\cite{dissanayake2020robust} & A & F & Pre-extracted feature representations (MFCCs/spectrogram); SHAP + Occlusion maps & Heart sound signals/UTS & L \& G  & Simple, stable and efficient architecture & Only applicable to DNN\\
\cite{schlegl2021scalable} & A & F & Explainable feature representations & Static MTS & G & Easy to visualize &  Weak interpretability due to the opacity of RNN-based AD; Fails to learn less frequent shapes\\
\hline  
\end{tabular}
\end{table}

\section{Literature Review on In-Model Techniques}

This section presents anomaly detection models that are considered to be inherently explainable. 
These anomaly detection models can provide insights into the relationships they have learned from the data, enabling an end-user to understand the decisions they have made. In general, the following methods are considered intrinsically explainable: 
\begin{itemize}
    \item Commonly seen transparent models in supervised learning, including Linear Models (Linear Regression, Logistic Regression), Decision Trees, Gaussian Process, Rule-based Learners, Generative Additive Models, and Bayesian Models;
    \item Feature subset based methods, including subspace anomaly detection methods, wrapper or embedded feature selection methods for anomaly detection;
    \item Miscellaneous other methods (mostly in an unsupervised setting) that reveal the rationale for how anomaly scores are calculated in a comprehensible way. 
\end{itemize}

\subsection{Transparent Models in Supervised Learning}

According to {\color{black}Lipton }\cite{lipton2018mythos}, a model is transparent if its intrinsic structure satisfies at least one of the following three requirements:
\begin{itemize}
    \item \textbf{Simulatability}: if a model can be simulated by a human, and thus whether it possible to reason about its entire decision-making process.
    \item \textbf{Decomposability}: if a model can be broken down into multiple parts, and these parts are easy to explain individually.
    \item \textbf{Algorithmic Transparency}: if a human can understand the process by which the model generates output from a given input.
\end{itemize}

In a supervised setting, commonly seen transparent models include Linear Models (such as Linear Regression and Logistic Regression), Decision Trees, Rule-based Learners in the form of \textit{if-then rules, m-of-n rules, list of rules, falling rule lists} or \textit{decision sets}, Gaussian Process, Generative Additive Models (GAMs), and Bayesian Models. Although anomaly detection is often an unsupervised problem, it can often leverage these methods in some way. However, transparency is not sufficient to guarantee explainability. Specifically, when a transparent model becomes exceedingly complex, it is not human-understandable anymore. Therefore, anomaly detection models that are developed based on these transparent models are considered to be explainable as long as they are not overly complex.

First, rule-based models are often leveraged to learn frequent patterns in the data, enabling interpretable anomaly detection. For instance, {\color{black}He et al. }\cite{he2005fp} apply frequent pattern mining to identify and explain anomalies in transaction data. Specifically, they leverage the Apriori algorithm \citep{agrawal1994fast} to find frequent patterns, and utilise the so-called top-$k$ contradictory frequent patterns to explain each identified anomaly. Similarly, {\color{black}Zhu et al. }\cite{zhu2012context} propose a model to capture frequent motion and background patterns of activities in video data, treating patterns that deviate from learned frequent patterns as anomalies. Likewise, {\color{black}Vacul{\'\i}k \& Popel{\'\i}nsk{\`y} }\cite{vaculik2016dgrminer} put forward the DRGMiner model, which mines frequent patterns in dynamic graphs and considers graphs deviating from these patterns as anomalous. Besides, {\color{black}Mauro et al. }\cite{mauro2017anomaly} propose HyVarRec to detect and explain anomalous traces for context-aware software product lines. Concretely, they apply Satisfiability Modulo Theories \citep{de2011satisfiability} to construct a conjunction of constraints that should be satisfied by normal traces when considering their contexts. As a result, a trace that violates the predefined constraints is considered anomalous. Moreover, {\color{black}B{\"o}hmer \& Rinderle-Ma }\cite{bohmer2020mining} develop the ADAR model to detect and explain anomalies in process runtime behavior. Specifically, ADAR leverages  association rule mining to extract a set of ordered rules that normal traces should satisfy. Hence, a test trace with a small support is considered anomalous. Importantly, they also propose a visualization technique called A\_Viz to show the rule violation. 

Second, decision trees and their variants have also been proposed to be used for the detection of anomalies, resulting in intrinsically explainable detection results. For instance, {\color{black}Kraiem et al. }\cite{kraiem2021human} introduce the \textit{Composition-based Decision Tree} (CDT) to detect and interpret anomalies in time series. Specifically, after preprocessing and labelling of given time series, a CDT is constructed as an extension of a decision tree on this labelled data, extracting rules for describing seen anomalies and detecting unseen anomalies. Also, the authors evaluate the explanation quality in terms of the number of used patterns and the length of rules. Furthermore, {\color{black}Cortes }\cite{Cortes2020ExplainableOD} presents an anomaly detection method that performs supervised decision tree splits on features, wherein the one-dimensional confidence intervals of each branch are built for the target feature. As a result, explanations can be obtained from the branching conditions and the general distribution statistics of non-anomalies that fall into the same branch. {\color{black}Besides, Aguilar et al. }\cite{aguilar2022towards} propose the Decision Tree-based AutoEncoder (DTAE) model to detect anomalies. Specifically, they use a decision tree to depict the encoding and decoding portions of AE, determining whether an instance is anomalous by comparing the input with the output. The advantage of using decision trees as encoders and decoders is that each tree contains the rules for categorizing tuples, offering interpretability. Meanwhile, {\color{black}Itani et al. }\cite{itani2020one} develop the so-called one-class decision tree (OC-Tree) model,  which employs Kernel Density Estimation to divide data subsets into intervals of interest and then encloses the data within hyperrectangles that can be explained by a set of rules. Additionally, {\color{black}Perez \& Lavalle }\cite{perez2011outlier} devise the alleged User Model to detect potential fraud in bank transactions, where they fit manually selected features into a threshold-based rule model, classifying the model outputs in the form of fraud probability into five categories.

Third, another line of research utilises regression models to perform anomaly detection, providing explanations for identified anomalies. For example, for each data instance,  {\color{black}Chen et al. }\cite{chen2012prediction} apply LOESS regression \citep{cleveland1979robust} by taking each feature in turn as the target variable and the remaining features as predictors based on its neighbours. An instance is considered anomalous in a certain feature if its actual value differs significantly from its predicted value. Particularly, for each identified anomaly, they provide a natural language explanation consisting of its considered neighbours and the associated feature differences. Besides, in {\color{black}Burak Gunay et al. } \cite{burak2019detection}, the heating and cooling load patterns of buildings are studied using three inverse models, including a univariate change point model, a regression trees based model, and an DNN based model. Particularly, change point models and regression trees are easy to interpret and can generate rules from their output. {\color{black}Moreover, Langone et al. }\cite{langone2020interpretable} leverage regularized Logistic Regression to identify anomalies in time series. In brief, they first utilise a bucket-based representation to represent the data, and then implement a rolling window procedure to extract features. On this basis, they employ the  Kolmogorov-Smirnov distance to select relevant features for anomaly detection, and the resulting features are fed to a Logistic Regression with Elastic Net regularization to detect anomalies. 

Fourth, some researchers utilize intrinsically interpretable models such as Gaussian Processes (GPs), Generalized Additive Models (GAMs), and Dynamic Bayesian Networks (DBNs) to detect anomalies. For instance, {\color{black}Berns et al. }\cite{berns2020towards} employ GPs to detect anomalies, where a GP is a stochastic process defined over a set a random variables such that every finite subset of these random variables follows a multivariate Gaussian distribution. If the actual value of a test instance deviates significantly from its predicted value, the GP model treats it as an anomaly. Meanwhile, {\color{black}Chang et al. }\cite{chang2022data} present an explainable anomaly detection model named DIAD based on GAMs. Specifically, a GAM model is a linear combination of smooth functions, where each function is defined on some variables. Given an anomaly, one can easily infer which features contribute the most to its outlyingness. Moreover, {\color{black}Slavic et al. }\cite{slavic2021interpretable} develop a DBNs based model to predict the state of a moving object in  Autonomous Driving domain, attempting to identify abnormal motion behaviors based on its motion direction and orthogonal direction. A test instance is considered anomalous if its predicted state  deviates significantly from its actual state. Due to the good properties of DBN, they can decompose the anomalous motion along its two directions and resort to the corresponding parameters to interpret the anomaly.

Finally, an important line of research attempts to introduce interpretable components in a complex anomaly detection model, providing weak interpretability. For instance, {\color{black}Zancato et al. }\cite{zancato2021stric} propose the STRIC model to detect anomalies in multivariate time series data. Specifically, STRIC consists of four layers. The first layer attempts to model the trend of time series by using a cascade of linear filters. The second layer implements a linear module to model and remove the seasonality at multiple time scales. Next, the third layer comprises a linear stationary filter bank that is able to approximate any trend or seasonality. Finally, the fourth non-linear layer consists of a randomly initialized Temporal Convolution Network model. Therefore, these four layers constitute a model capable of predicting time series. On this basis, they extend the CUMSUM algorithm \citep{yashchin1993performance} to detect anomalies by using the likelihood ratio between two windows of prediction residuals. Particularly, the linear components used in STRIC provide interpretability.

\textit{Discussion:} To take advantage of these transparent models for anomaly detection, one usually needs to turn the unsupervised anomaly detection problem into a supervised or semi-supervised setting. For instance, References \cite{vaculik2016dgrminer,bohmer2020mining,slavic2021interpretable} attempt to learn normal behaviours or patterns  by training the model on exclusively normal data, and then identify anomalies by comparing a test instance with the expected normal behaviours. Meanwhile, References \cite{kraiem2021human,chen2012prediction} either directly leverage labelled data or decompose an unsupervised problem into many supervised problems \citep{paulheim2015decomposition}.

\subsection{Feature Subset Based Models} \label{Sec:InModel2}
The methods in this subsection select one or more subsets of features to detect and explain anomalies. Specifically, it contains subspace anomaly detection methods and feature selection methods for anomaly detection. Subspace anomaly detection methods find anomalies that are only detectable in certain subspaces, providing intrinsic explanations based on subspaces. Moreover, wrapper or embedded feature selection methods select a subset of original features that are relevant for anomaly detection, thereby improving the interpretability of detection results. Note that wrapper or embedded feature selection methods select features during the process of performing anomaly detection, not before the anomaly detection process (see Subsection~\ref{Sec:PreModel1}). Furthermore, feature selection methods can be considered as a special case of subspace anomaly detection since they actually select a subspace for anomaly detection. 

First, subspace anomaly detection usually consists of two steps: finding subspaces and assigning anomaly scores. Subspace anomaly detection has received extensive attention, resulting in a collection of strategies for finding subspaces and assigning anomaly scores. In general, finding subspaces and assigning anomaly scores can be decoupled or intertwined.  For instance, {\color{black}Muller et al. }\cite{muller2011statistical} propose OUTRES. For each instance, they first use the Kolmogorov-Smirnov goodness of fit test to exclude some subspaces from the powerset of features. Specifically, they exclude subspaces in which the local densities of the given instance and its neighbourhood are uniformly random distributed. Then, for the residual subspaces, they define a dimensionality-unbiased anomaly scoring function to measure the local density deviation of the given instance. Next they aggregate the anomaly scores of each instance across its non-uniformly random distributed subspaces. Moreover, {\color{black}Keller et al. }\cite{keller2012hics} present HICS. Specifically, they first look for subspaces with high contrast by measuring the correlation between features in a subspace using statistical tests, viz. the difference between  marginal probability density and conditional probability density. Second, they apply an off-the-shelf anomaly ranking method such as LOF \citep{breunig2000lof}  on selected subspaces and aggregate the results. It can be seen that both methods  \cite{muller2011statistical, keller2012hics}  decouple subspace search and anomaly scoring. In contrast, {\color{black}Dang et al. }\cite{dang2014discriminative} leverage spectral graph theory to achieve subspace anomaly detection. Specifically, they first construct an undirected graph that can capture the local geometry of all instances. Second, they attempt to learn an optimal subspace that can separate well an instance from its neighbors, while preserving the intrinsic geometrical structure of data. Correspondingly, a well separated instance is considered anomalous and the corresponding subspace acts as an explanation. The subspace search and anomaly scoring are intertwined in this method.

Some researchers attempt to leverage dimensionality reduction or feature projection techniques to perform subspace anomaly detection. For example, given a data instance with its global nearest neighbors,  {\color{black}Kriegel et al. }\cite{kriegel2012outlier} first project these data instances with all $d$ features into subspaces of varying size, where the subspace is spanned by the $l$ largest principle components using robust PCA. Meanwhile, they compute the projection to the subspaces spanned by the remaining $d-l$ principle components as its error vectors. Second, they choose the error vector with the largest $L_{2}$-norm value as its anomaly score and explanation. The rationale of using PCA is that the correlation dimensionality is highly related to the intrinsic dimensionality of data. Meanwhile, {\color{black}Bin et al. }\cite{bin2016abnormal} develop the ASPCA model. Given a dataset with $D$ features, they first compute and order the principal components using sparse PCA \citep{jolliffe2003modified}. The first $k$  principle components which capture most of the variance are called \textit{normal subspace}, and the remaining $D-k$ components are called \textit{abnormal subspace}. An instance is considered anomalous if its has large projection length in the \textit{abnormal subspace}. Based on sparse PCA, each feature is a linear combination of a few original features. Therefore, this method can easily obtain the original features that are responsible for an anomaly, resulting in an explanation. Furthermore, {\color{black}Dang et al. }\cite{dang2013local} introduce the LODI model. For each instance, they first select its neighbors using an information-theoretic tool. Second, they use local dimensionality reduction to select an optimal subspace in which this instance can be maximally separated from its neighbours. An instance that is relatively easy to separate is considered anomalous. More concretely, the local dimensionality reduction problem is solved via matrix eigen-decomposition, which can return the corresponding original features that are most important to explain an anomaly. Additionally, {\color{black}Pevn{\`y} }\cite{pevny2016loda} presents Loda, an online anomaly detection model that can also provide explanations. Specifically, Loda first leverages sparse random projections to obtain a collection of one-dimensional subspaces. Second, it constructs a histogram in each subspace, aiming to approximate the probability density. Third, it aggregates these one-dimensional histograms to estimate the joint probability density. Consequently, an instance with low estimated  probability density is considered anomalous. For each identified anomaly, Loda can rank features according to their contributions to the anomaly score as an explanation. 

As we can see, the above mentioned methods utilise some well-defined criteria to search subspace and then assign anomaly scores. However, another line of research intends to use random search strategies to search for subspaces. For instance, {\color{black}Keller et al. }\cite{keller2013flexible} propose RefOut, which consists of three steps. They first generate an initial subspace pool that is a set of randomly selected subspaces. On this basis, they utilise an off-the-shelf anomaly detection model to perform anomaly detection, resulting in a set of anomaly scores for each instance. Second, for each instance, they generate a refined subspace by maximizing the discrepancy of anomaly scores. Aggregating all these refined subspaces leads to a refined subspace pool. Third, they apply again the anomaly detection model on the refined subspace pool to obtain an anomaly score for each instance. Considering that the cardinality of each refined subspace may be different, they  normalize the anomaly scores to ensure comparability. Accordingly, they return the maximum anomaly score and the corresponding subspace for each instance as an explanation. {\color{black}Similarly, Savkli \& Schwartz }\cite{savkli2021random} put forward RSMM. Concretely, they first randomly select $m$ subspaces of dimension $k$, ensuring that each dimension contributes equally to the final probability model. Second, they construct a mixture model such as Gaussian Mixture Model in each subspace. Third, they compute the geometric averaging of the probability densities in all subspaces as the joint probability density. Therefore, if a test instance is located in a low-density region, it is considered anomalous. Furthermore, to interpret an anomaly, they rank features according to how often they consist in the subspaces where the anomaly is considered an anomaly.

Second, despite the prevalence of subspace anomaly detection, other techniques such as feature selection have also been exploited to facilitate the interpretability of anomaly detection. For instance, 
{\color{black}Pang et al. }\cite{pang2017learning} introduce a wrapper feature selection framework for anomaly detection. Specifically, they first create an internal evaluation metric for anomaly detection and then select relevant features for detecting anomalies by iteratively maximizing this metric. They have only applied this framework on their proposed anomaly detection model though, which only works on categorical data. {\color{black}Besides, Pang et al. }\cite{pang2018sparse} propose an embedded feature selection method for anomaly detection, dubbed CINFO, which is an ensemble of sequential ensemble learners. Specifically, the base learner, namely the sequential ensemble learner, iteratively and mutually refines the anomaly detection and feature selection processes. In this way, they build many similar base learners, which are then aggregated to produce the final anomaly scores. Hence, the method does not explicitly provide any selected features and lacks interpretability due to the use of an ensemble approach. Meanwhile, {\color{black}Roshan \& Zafar }\cite{roshan2021utilizing} develop an AutoEncoder (AE) based model incorporating the SHAP technique to detect and explain anomalies in computer network data. Specifically, they first train an AE model on exclusively normal computer network data with all input features, followed by applying Kernel SHAP to explain the predictions of the trained AE model. Next, they use the trained AE model to detect anomalies in another dataset containing cyberattacks, and then apply again Kernel SHAP to explain the predictions, aiming to select a subset of important features to identify anomalies. Finally, these selected features are used to train a refined AE model for anomaly detection.

\textit{Discussion:} The subspace anomaly detection methods introduced here are considered inherently explainable since they only explain anomalies identified by themselves. In other words, the \textit{Detection-Definition} and \textit{Explanation-Definition} of anomaly are usually consistent. Therefore, the generated explanations are intrinsic regardless of whether the anomaly detection and subspace search processes are interleaved or decoupled. In contrast, the subspace anomaly detection methods that will be presented in the \textit{shallow post-model techniques} section are distinct, as they aim to interpret anomalies that are identified by other detection models or experts. As a result, the \textit{Detection-Definition} and \textit{Explanation-Definition} of an anomaly are likely to be different since the \textit{Detection-Definition} is generally unknown.  

\subsection{Other Miscellaneous Models in Unsupervised Learning}

In principle, an anomaly detection model that reveals the rationale for how anomaly scores are calculated in a human comprehensible way can be considered intrinsically explainable. Hereinafter, we survey a collection of intrinsically explainable anomaly detection methods that do not belong to the commonly seen transparent methods in supervised learning nor feature subset based methods. Due to the diversity of these methods, i.e., they share few basic techniques, we organize them according to the type of data they have been designed for.

\subsubsection{Models for Tabular Data}
A plethora of models have been devised to detect anomalies in tabular data whilst providing intrinsic explanations. For instance, as a typical method, distribution based anomaly detection models attempt to fit data with probabilistic distributions. Then, data instances that do not conform to the fitted model are considered anomalous. {\color{black}According to Agyemang et al. \cite{agyemang2006comprehensive},} distribution based anomaly detection techniques are intrinsically explainable. This is because the identified anomalies can be meaningfully interpreted from a statistical perspective once the probabilistic distribution is known. {\color{black}Dunstan et al. }\cite{dunstan2009anomalies} {\color{black}take a different approach and} utilise a data cube structure to divide transaction data instances into different regions. They refer to each region as a context and show how each instance can be abnormal in different contexts. More importantly, they create anomaly tables and anomaly lattices to explain anomalies. An anomaly table contains anomalous transactions alongside their contexts. Meanwhile, an anomaly lattice graphically displays the anomalies with their contexts. {\color{black}Rather than using groups to define contexts in which anomalies can be detected, Mejia }\cite{mejia2010outlier} adapts Adaptive Resonance Theory (ART) \citep{carpenter1987neural} to group instances into clusters such that instances residing in the smallest clusters are considered anomalous. By virtue of the good properties of ART, one can obtain the feature differences between every two clusters, resulting in explanations for the anomalous instances. 

{\color{black}Smets \& Vreeken }\cite{smets2011odd} {\color{black}take a more global approach to anomaly detection, i.e., they} employ the Minimum Description Length (MDL) principle to determine whether a data instance is anomalous; in brief the number of bits required to encode it using compression is used as anomaly score. They utilise the Krimp algorithm \citep{siebes2006item} as the compressor, which is trained on exclusively normal samples to capture normal behaviours. On this basis, they provide explanations by showing which patterns were recognised in the anomalies, as well as by checking whether small changes can turn the anomalies into normal instances. If it can, the anomalies are observation errors rather than real anomalies.
{\color{black}Instead of using patterns to represent what is normal, Park \& Kim }\cite{park2021explainable} put forward  a model that is an ensemble of Region-Partition (RP) trees. Each RP tree is trained only on normal data and thus represents a partition of the normal data region. Hence, if a test instance can arrive at a leaf node of any individual RP tree, it is considered normal. Otherwise, it is an anomaly. Considering that the size of each RP tree is small, one can easily find the hypercube in which the anomaly is stuck. On this basis, they take the intersection of hypercubes of all RP trees as an explanation for the anomaly. 

While the above mentioned models focus on static tabular data, {\color{black}Dickens et al. }\cite{dickens2020interpretable} propose Mondrian Pólya Forest (MPF) to detect and explain anomalies on large data streams by combining random trees with non-parametric density estimation approaches.  Specifically, the Mondrian Process \citep{roy2008mondrian} is a family of hierarchical binary partitions of data and the Pólya Tree \citep{mauldin1992polya} is a non-parametric approach that can estimate the density function of binary partitions. They combine the Pólya Tree structure with a truncated Mondrian Process to deal with static data, and combine the Pólya Tree structure with a Mondrian Tree to handle data streams. In this way, they construct a forest, namely MPF, for density estimation and anomaly detection. As a result, an instance with relatively low estimated density is considered anomalous. Furthermore, with the good properties of MPF, the resulting anomaly scores are probabilistic and therefore interpretable. 
 
\subsubsection{Models for Sequential Data}

{\color{black}One line of research addresses the problem of detecting and providing intrinsic explanations in time series data, which is a type of sequential data.} Techniques such as sparse learning and  time series decomposition have been leveraged to develop intrinsically interpretable anomaly detection models for time series data. For instance, {\color{black}Li et al. }\cite{li2021stacking} apply deep Generative Models (DGMs) to detect anomalies in multivariate time series data. Specifically, they first set up a stacking Variational AutoEncoder (VAE) based model that constructs a single-channel block-wise reconstruction, followed by stacking it multiple times using a weight sharing technique to handle channel-level similarities. Second, they utilise a graph learning module to learn a sparse adjacency matrix for every channel, attempting to extract structure information for achieving an explainable reconstruction process. As a result, a test instance with a large reconstruction error is considered anomalous. Meanwhile, {\color{black}Cheng et al. }\cite{cheng2021multi} exploit time series decomposition techniques from a multi-scale perspective to identify spatiotemporal abnormalities of human activity. They first employ the Seasonal-Trend decomposition with the Loess (STL) method to decompose the time series to look for anomalies. As a result, the periodic as well as trend components of observed data are eliminated during the time series decomposition, and the remaining components capture anomalous activity signatures. By examining the residual elements of the time series for each spatial unit, they are able to identify spatiotemporal anomalies in human activity. Finally, they devise a rule to match anomalies identified at different scales in accordance with their spatiotemporal influence ranges and explain anomalies based on their multi-scale characteristics. 

{\color{black}Another line of research focuses on the task of identifying and offering intrinsic explanations in network traffic data, which can be seen as an instance of other sequential data.} For example, {\color{black}Grov et al. }\cite{grov2019towards} first group the network traffic data into different sessions, followed by learning two behavioral models, namely a Markov Chain (MC) model and a Finite State Automata (FSA) model, on normal sessions. Next, for an incoming session, they compute a similarity measure with respect to these two models, resulting in an anomaly score. More concretely, the MC model returns two probabilities as the anomaly score and the FSA model returns a distance as the anomaly score. Meanwhile, {\color{black}Mulinka et al. }\cite{mulinka2020human} present HUMAN, a hierarchical clustering based method. Specifically, they consider three different clustering methods to group data instances into clusters, assuming that the normal behaviour is represented by the largest cluster. Therefore, data instances residing in the smallest clusters are considered anomalous. Next, they explain the detected anomalies by displaying the clustering results, including the number of clusters, the size of each cluster, and a textual explanation of each cluster. Moreover, {\color{black}Marino et al. }\cite{marino2022self} propose the Network Transformer model (NeT). First, the network data is represented by a graph where its nodes represent network device IP addresses and the edges describe data packets delivered between different devices. Second, NeT extracts hierarchical features from the graph for anomaly detection. Third,  based on these features, NeT employs existing anomaly detection models---such as LOF \citep{breunig2000lof}, OCSVM \citep{scholkopf1999support}, and AutoEncoders---to identify anomalies at various granularity levels. Moreover, NeT provides explanations based on the graph structure, offering a subset of hierarchical features that allow users to pinpoint the devices affected by the anomalies and the connections that caused the anomalies.

\subsubsection{Discussion} 

Due to the lack of a unified definition of anomaly and the diversity of data types, a wide range of in-model XAD techniques have been explored in an unsupervised setting. More concretely, techniques such as Probabilistic Models (e.g., Mondrian Pólya Forest and other distribution or density estimation based approaches), Data Cube structure, Incremental Clustering, MDL-based Pattern Compression, Region-Partition trees are harnessed to detect and explain anomalies in tabular data. Meanwhile, techniques such as Markov Chain, Finite State Automata, Hierarchical Clustering, Sparse Learning in VAE, Time Series Decomposition, and Hierarchical Features in Graph Representation are adapted to identify and interpret anomalies in sequential data.

\subsection{Summary}
As shown in Table~\ref{tab:summary_intrinsic}, the in-model techniques presented in this section are model-specific. Moreover, the majority of these methods provide feature-based explanations (including pattern-based explanations), with the exception of References \cite{he2005fp,grov2019towards,cheng2021multi}, which offer sample-based explanations, and References \citep{chen2012prediction,dang2013local} generate explanations from both perspectives. 

According to their main characteristics, we subdivide them into three high-level groups, i.e., \textit{transparent models in supervised learning, feature subset based models}, and \textit{miscellaneous models in unsupervised learning}, for which we make the following observations.

\begin{landscape}
\small
\begin{longtable}{p{0.03\textwidth}p{0.02\textwidth}p{0.05\textwidth}p{0.23\textwidth}p{0.11\textwidth}p{0.03\textwidth}p{0.32\textwidth}p{0.33\textwidth}
}
\caption{Summary of in-model XAD techniques. \textit{Spec} indicates whether a method is model-agnostic (A) or model-specific (S). Note that all in-model techniques are basically model-specific. \textit{Pers} specifies whether a method is feature-based (F), sample-based (S) or pattern-based (P). \textit{Tech} indicates the techniques used in each method. \textit{Data} indicates the data type for which the method is applicable (TN: Tabular Numeric; TC: Tabular Categorical; TM: Tabular Mixed; UTS: Univariate Time Series; MTS: Multivariate Time Series; ES: Event Sequence). \textit{Loc} shows whether a method provides a local explanation (L) or global explanation (G). Moreover, \textit{Pros} and \textit{Cons} describe the advantages and disadvantages of each method, respectively.}
\label{tab:summary_intrinsic}\\
\hline Ref & Spec & Pers & Tech & Data &Loc  & Pros & Cons\\
\hline
\cite{he2005fp} & S & P & Frequent pattern mining & Static TC & G & Performs frequent pattern mining and outlier discovery simultaneously & Only applicable to categorical data \\
\cite{perez2011outlier} & S & S & Rule-based model to produce probability & Static TN & G & Fast & Low accuracy \\
\cite{zhu2012context} & S & P & Frequent pattern mining & Static video & G & Able to detect point anomalies, contextual anomalies, and collective anomalies & Needs labeled normal activity data\\
\cite{vaculik2016dgrminer} & S & P & Frequent pattern mining  & Streaming dynamic graph & G & Able to handle dynamic graphs & Computationally expensive \\
\cite{mauro2017anomaly} & S & P & Satisfiability Modulo Theories based rule models & Static ES & G & Integrates contexts to detect anomalies & Computational expensive; Explanations may be too long to understand \\
\cite{bohmer2020mining} & S & P & Association rule mining + Visualisation & Event logs & L & Able to handle process change and flexible executions; Evaluating the quality of explanations & Assumes the availability of domains, process models, and anomalies in evaluation \\
\cite{kraiem2021human} & S & F & Composition-based decision tree & Static UTS & G & No manual tuning of hyper-parameters; Evaluating the quality of explanations & Needs labeled data \\
\cite{itani2020one} & S & F & One Class Decision Tree & Static tabular & G & Produces compact and readable results  &  Parameterization of the KDE is difficult\\
\cite{aguilar2022towards} & S & F & Decision tree based AE & Static TC & G & Intrinsically interpretable AE & Not suitable for dataset with many features; Only applicable to categorical data \\
\cite{chen2012prediction} & S & F \& S & LOESS regression & Static TM  & G & Able to handle heterogeneous features; Evaluating the quality of explanations & Sensitivity to important parameters not discussed\\
\cite{burak2019detection} & S  & F & Change Point Model + Regression Trees & static tabular & G & More insights from multiple models  & No explanations for ANN  \\
\cite{langone2020interpretable} & S  & F & Logistic regression & Streaming MTS & G & Able to predict short-term anomalies & Not able to predict long-term anomalies  \\
\cite{berns2020towards} & S  & F & Gaussian Process & Streaming TN  & G & Able to handle unreliable, noisy, or partially missing
data & High computation cost; Hard to do model selection; Poor at handling discontinuities  \\
\cite{slavic2021interpretable} & S & F & Dynamic Bayesian Networks & Streaming MTS & G & Able to incorporate domain knowledge & Computationally expensive \\
\cite{zancato2021stric} & S & F & Linear components & Static MTS & G & End-to-end training & Weak interpretability due to deep models \\
\cite{chang2022data} & S  & F & Generalized Additive Models & Static TM & G & Able to incorporate a small amount of labeled data & Relies on assumptions about the data generating mechanism \\
\cite{muller2011statistical} & S & F & Subspace AD & Static TN  & L & More scalable than other density-based methods & Weak interpretability due to ensemble\\
\cite{keller2012hics} & S & F & Subspace AD & Static TN & L & Adjustable to any AD & Weak interpretability due to ensemble\\
\cite{kriegel2012outlier} & S & F & PCA + Subspace AD & Static TN  & L & Works with arbitrarily oriented subspaces & Weak interpretability by using error vectors of PCA; Does not work well with high-dimensional data\\
\cite{bin2016abnormal} & S & F & Sparse PCA + Subspace AD  & Static TN  & G  & Fast & Non-trivial parameters setting by end-users \\
\cite{dang2013local} & S & F \& S & Subspace AD & Static TN & L & Explores interconnections between neighboring members & High computational cost; Assumes linear separability of anomalies with their neighbours\\
\cite{keller2013flexible} & S & F & Subspace AD & Static tabular & L & Applicable to any AD model & Computationally expensive\\
\cite{dang2014discriminative} & S & F & Subspace AD & Static TN  & L & Ensures quality of explanation via keeping local geometry & Not scalable to high-dimensional data\\
\cite{savkli2021random} & S & F & Subspace AD & Static TM & L & Applicable to mixed data; Parallelizable; Applicable to high-dimensional data &  Difficult to initialize clusters for GMMs in high-dimensional subspaces\\
\cite{pevny2016loda} & S & F & Feature projections & Streaming and static TN  & L & Fast; Adapted to concept drift; Able to handle missing variables & Only considers one-dimensional projections; Weak interpretability due to ensemble \\
\cite{pang2017learning} & S & F & Wrapper feature selection & Static TC & L \& G & Works well with noisy features & Only applicable to categorical data\\
\cite{pang2018sparse} & S & F  & Embedded feature selection  & Static TC & L  & Works with high-dimensional data & Lacks interpretability due to the use of an ensemble approach\\
\cite{roshan2021utilizing} & S & F & SHAP based feature selection  & Static MTS/ES  & G & Does not need labelled data & Only applicable to AE-based model; High time complexity with kernel SHAP \\
\cite{agyemang2006comprehensive} & S & F & Statistical distribution & Static tabular   & L \& G & Sound statistical foundation & Assumes probabilistic distribution of data \\
\cite{dunstan2009anomalies} & S & F & Data cube &  Static tabular & L & Considers anomalies in multiple contexts & Computationally expensive \\
\cite{mejia2010outlier} & S & F & Incremental clustering & Static TM   & G & Fast & Non-trivial setting of the threshold parameter \\
\cite{smets2011odd} & S & P & MDL-based pattern compression  & Static TC  & L & Provides detailed inspection and characterisation of decisions & Only applicable to categorical data \\
\cite{grov2019towards} & S & P & Markov Chain; Finite State Automata & Network traffic data & G & Robust to new unseen anomalies & Trains on normal data \\
\cite{mulinka2020human} & S & S & Clustering   & Network traffic data  & G & Able to integrate domain knowledge & Non-trivial setting of parameters \\
\cite{dickens2020interpretable} & S & F & Bayesian nonparametric based density estimation & Static and steaming TM  & G & Able to handle streaming data  & Density estimation does not work well in high-dimensional data \\
\cite{park2021explainable} & S & F & RP Tree  & Static tabular   & G & Automatically determines the threshold on anomaly scores & Weak interpretability due to ensemble \\
\cite{li2021stacking} & S & F & Sparse learning + Reconstruction error  & Static MTS  & G & Considers single-channel anomalies and  structural multi-channel anomalies & Only considers linear correlation between channels; Weak interpretability due to VAE \\
\cite{cheng2021multi} & S & S & Time series decomposition & Spatial-temporal data & G &  Considers multi-scales to gain more insights  into anomalies & Does not work well with data missing and deficiency \\
\cite{marino2022self} & S & F & Hierarchical features in graph  & Static graph  & G & Self-supervised training that does not need labeled data & Weak interpretability due to deep models \\
\hline  
\end{longtable}
\end{landscape}
%\end{table}
First,  \textit{for transparent models in supervised learning}, we find that most methods can provide global explanations as their entire logic can be easily understood by humans due to their transparent nature. However, \textit{decision tree based models} are hard to explain when the tree is too deep or too wide. To alleviate this problem, feature selection can be leveraged. Meanwhile, an ensemble of decision trees can avoid overfitting of the data, thereby improving the generalization performance. However, an ensemble of trees is not human-understandable. Concerning  \textit{rule-based models}, a large set of rules or a long rule is difficult to explain. Therefore, a human-reasonable size is required to maintain interpretability. Furthermore, an inherent problem of using \textit{linear models} for interpretation is that when the model does not fit the training data optimally, it may optimize errors using spurious features that may be difficult to interpret for humans \citep{guidotti2018survey}. Overall, for these transparent models to retain their interpretability characteristics, they must be limited in size and the features used should be understandable to the end-users \citep{belle2021principles}.

Second, for \textit{feature subset based models}, most methods can only provide local interpretations, that is, only a certain output can be interpreted at a time. Besides, some subspace anomaly detection methods do not provide explicit explanations due to the use of ensemble techniques to aggregate anomaly scores in multiple subspaces. However, the contribution of each feature is relatively easy to obtain.  Moreover, most subspace anomaly detection methods were originally designed to tackle the issue of \textit{curse of dimensionality} when detecting anomalies in high-dimensional data \citep{zimek2012survey}. Therefore, promoting interpretability is not their main concern. Notably, we observe that there is extremely limited research on wrapper or embedded feature selection for anomaly detection. 

Third, for \textit{miscellaneous models in unsupervised learning}, these methods are quite different from each other, as they are specifically designed for the anomaly detection and not explored in a supervised setting. Given the lack of a unified definition of anomalies and the diversity of data types, it is not surprising that these approaches are very diverse. Importantly, we note that most of these methods can provide global explanations, in the sense that the logic of the whole model is human-understandable, or some important properties of the model can be leveraged to interpret all decisions.  

\section{Literature Review on Shallow Post-Model Techniques}

\textit{Post-model} methods inspect an anomaly detection model after the detection process is completed, or just inspect a given anomaly without being given an anomaly detection model. In other words, these techniques do not interfere with the anomaly detection process, operating only on the basis of correlating the input of the anomaly detection model (if any) with its output. Due to the proliferation of techniques in this category, in this section we only introduce techniques designed for non-deep learning processes, and we call them \textit{shallow post-model anomaly explanation} techniques.

Most \textit{shallow post-model anomaly explanation} methods intend to find a subspace or a set of subspaces in which the given anomaly differs the most from other instances, and we call these methods \textit{subspace based methods}. \textit{Surrogate methods}, on the other hand, resort to identify another model to explain the anomaly detection model or just the given anomalies. Specifically, a surrogate model can be a transparent model, such as a set of rules or a decision tree, or an opaque model, such as XGBoost or SVM. Importantly, if the surrogate model is an opaque model, it should be easy to interpret by using XAI techniques such as SHAP. Meanwhile, \textit{miscellaneous methods} such as comparing patterns to find differences, leveraging SHAP techniques to measure feature importance, and visualisation also play an important role in shallow anomaly explanation.

\subsection{Subspace based methods}
%An introduction of subspace based methods and illustrate its difference with in-model subspace anomaly detection

Given an anomaly or a group of anomalies, \textit{subspace based methods} aim to find a subspace or a set of subspaces in which the anomaly deviates the most. Different from the intrinsically explainable \textit{subspace anomaly detection} methods that were introduced in Section \ref{Sec:InModel2}, the subspace based methods investigated hereinafter do not assume the availability of anomaly detection models. 

First, different explanation methods usually have different definitions for anomaly, dubbed \textit{Explanation-Definition} in this survey, leading to different measures of abnormality. For instance,  {\color{black}Knorr \& Ng }\cite{knorr1999finding} define strongest and weak outliers, based on which they use so-called intensional knowledge to explain anomalies. For each anomaly identified in the original feature space, they report the minimal subspaces in which it behaves anomalously. Particularly, in their proposed algorithm CELL, for each instance, they utilize the number of neighbors in its local neighborhood of a given radius as the anomaly score. However, this anomaly measure can be replaced by other anomaly measures such as density, depth, etc. As far as we know, their work is seminal in anomaly interpretation. Additionally, {\color{black}Zhang et al. }\cite{zhang2004hos} propose HOS-Miner to identify outlying subspaces for a given anomaly. Specifically, they define the sum of distances between the anomaly and its $k$-nearest neighbors as its anomaly score in each subspace, thereby returning the outlying subspace with the lowest dimensionality as an explanation. {\color{black}Similarly, Micenkov{\'a} et al. } \cite{micenkova2013explaining} propose an anomaly explanation technique that works on tabular dataset with numeric features. Specifically, given an anomaly, they look for a subspace in which this instance is well separable from the rest. To achieve this, they first generate a classification dataset consisting of a comparable number of normal and abnormal instances. Second, they apply an existing feature selection method to find a subset of features that are relevant for the classification, namely separation. Particularly, they define a measure of separability based on the probability density function of a normal distribution as the anomaly measure. Finally, the obtained subspace serves as an explanation for the anomaly. 

Second, some researchers attempt to provide explanations from multiple perspectives or contexts. For instance, {\color{black}Angiulli et al. }\cite{angiulli2009detecting} propose a method that is capable of providing explanations from both global and local perspectives. On the one hand, for an anomaly, they measure its abnormality with reference to all data instances in different subspaces, delivering the subspace with the highest abnormality as a global explanation. On the other hand, for an anomaly, they first select a subset of features and the corresponding values to define a reference group, and then compute its abnormality with respect to this reference group in different subspaces. Accordingly, the reference group and the corresponding subspace with the highest abnormality constitute a local explanation. Note that the definitions of global explanation and local explanation used in Reference \cite{angiulli2009detecting} differ from those we defined in this survey. Particularly,  the \textit{abnormality} is defined in terms of the frequency of the anomalous instance and the frequencies of referencing instances. {\color{black}Furthermore, M{\"u}ller et al. }\cite{muller2012outrules} present OutRules, which generates multiple explanations for an anomaly in different contexts. Specifically, OutRules explains an anomaly by generating rules that describe the deviation of this instance in contrast to its context. On the one hand, a subset of features are used to define a context consisting of highly clustered instances. On the other hand, they attempt to find an extended subset of features in which one of these instances is significantly deviating. Concretely, the anomaly measure used in their framework can be instantiated by the underlying anomaly score of any anomaly detection model such as LOF. Similarly, {\color{black}Angiulli et al. }\cite{angiulli2017outlying} devise a method that consists of two steps. Given an anomaly and a dataset, for each feature, they first determine the interval that includes the anomaly and the associated condition, resulting in a set of conditions on all features. Second, they employ an Apriori-like strategy to search for \textit{explanation-property pairs} for the anomaly. More concretely, an \textit{explanation} is a set of conditions used to define a context where the anomaly is located. Meanwhile, a \textit{property} is an additional condition posed on a feature other than the features that are used to define the context, aiming to distinguish the anomaly from the context. In particular, they define an anomaly measure based on the probability density function of each feature. Consequently, the \textit{explanation-property} pair and the corresponding anomaly score constitute an explanation for the anomaly.

Third, other techniques such as visualisation can be leveraged to further improve the explainability of subspace based methods. For instance, given a dataset consisting of real-valued features and a list of anomalies, {\color{black}Gupta et al. }\cite{gupta2018beyond} propose the so-called LOOKOUT approach to explain these anomalies. Specifically, they attempt to find a limited number of 2-dimensional subspaces in which the given anomaly deviates the most from the rest. Particularly, if the anomaly is detected by an anomaly detection model, they utilise the underlying anomaly measure to obtain the anomaly score. Otherwise, they employ any other off-the-shelf model such as LOF to obtain the anomaly score. Moreover, they visualise these subspaces using 2-dimensional scatter plots, known as focus plots in their paper, and then present these \textit{focus plots} to the end-users as explanations. Also, the above-mentioned approach OutRules \cite{muller2012outrules} utilises parallel coordinates plots to visualise the anomalies.

Fourth, some methods have been explored to explain anomalies in a group. For instance, {\color{black}Angiulli et al. }\cite{angiulli2012discovering} extend their previous work \citep{angiulli2009detecting} to explain a group of anomalies. Furthermore, {\color{black}Macha \& Akoglu }\cite{macha2018explaining} propose x-PACS to explain anomalies in a group in three steps. They first {\color{black} utilise a subspace} clustering algorithm to identify clusters that the anomalies form. Second, for each subspace cluster of anomalies, they leverage an axis-aligned hyper-ellipsoid to represent it. Third, they employ the MDL criterion to identify a set of hyper-ellipsoids that are compact, non-redundant, and pure. Particularly, x-PACS does not require a measure of outlyingness since they address the anomaly explanation problem from a subspace clustering perspective. 

Fifth, with the emergence of many anomaly explanation methods, some researchers endeavour to formally define the anomaly explanation problem or propose a taxonomy of existing methods. Concretely, {\color{black}Kuo \& Davidson }\cite{kuo2016framework} formally define the outlier description (namely anomaly explanation) problem and  propose a Constraint Programming (CP) based framework to encode the problem. Particularly, they utilise a neighborhood density based criterion to measure the outlyingness of an instance in each subspace. On this basis, they introduce a CP framework to learn the optimal subspace to explain an anomaly. Their framework and variants can explain an anomaly in a single subspace, multiple subspaces, or by introducing the human in the loop. Meanwhile, {\color{black}Vinh et al. }\cite{vinh2016discovering} for the first time divide outlying aspects mining techniques into two categories, viz. feature selection based approaches and score-and-search approaches, and additionally make two important contributions. First, they formalize the concept of dimensionality-unbiasedness for anomaly scoring functions. They show that some widely used anomaly scoring functions such as distanced-based and density-based scoring measurements violate this important property. Moreover, they put forward two dimensionality-unbiased anomaly scoring functions, namely Z-score and isolation path score, to measure the outlyingness of an instance in different subspaces. Second, they propose a beam search framework to overcome the limitation of exhaustive search in exponentially large space. Consequently, for an instance, they return the subspace with the highest dimensionality-unbiased anomaly score as an explanation. However, {\color{black}Samariya et al. }\cite{samariya2020new} point out an issue of using Z-score normalisation of density to rank subspaces for outlying aspects mining. Particularly, Z-score normalisation has a bias towards data distribution (with high variance subspaces) although it is dimensionality-unbiased. To tackle this issue, they propose another anomaly scoring function called SiNNE for outlying aspects mining. Specifically, SiNNE consists of an ensemble of models where each model is developed based on a subset of data. However, due to the use of ensemble techniques, SiNNE cannot provide explanations for identified anomalies. 

Finally, another line of research attempts to explain any given instance that may be anomalous or normal. For example, given a data instance, {\color{black}Duan et al. }\cite{duan2015mining} develop a method to find its minimal outlying subspace, i.e., the subspace with the lowest dimensionality where the query instance is most deviating from others. To achieve this, they first assume that the instances are generated from a probability distribution that is often unknown. Second, they utilise kernel density estimation techniques to approximate the probability density of an instance in each subspace, deriving its anomaly score. Moreover, they employ heuristic techniques to prune the set of possible subspaces that need to be explored. 

\textit{Discussion:} Most subspace based methods reviewed above suffer from two limitations: high computational costs and poor explanation fidelity. First, most methods intend to find a minimal subspace in which the anomaly deviates the most. To find such a subspace, they usually need to go through the exponentially large search space. Although some pruning techniques such as beam search are leveraged to mitigate this problem, optimality is no longer guaranteed. Second, almost all methods in this category have their own definitions of anomaly (namely \textit{Explanation-Definition}) when trying to interpret anomalous instances. Importantly, this \textit{Explanation-Definition} is very likely to  differ from the \textit{Detection-Definition}, leading to a poor fidelity of the explanation. In other words, if the anomaly detection model is available, the provided explanation may not reflect its actual decision-making process.

\subsection{Surrogate methods}
%An introduction of surrogate models, what it means, which typical models it contains, etc.

A line of research in shallow post-model XAD techniques is to utilise surrogate models to describe given anomalies or anomaly detection models. In general, the surrogate model can be a transparent model or an opaque model. If a transparent model such as a set of rules or a decision tree is employed to depict the anomaly, the result is directly understandable. However, if an opaque model such as XGboost or SVM is leveraged to approximate the outputs, additional XAI techniques such as SHAP or LIME are required to make the results understandable.

First of all, model-agnostic rule learners are often leveraged to  extract a set of rules or patterns as the surrogate model, aiming to explain anomalies. For instance, {\color{black}Ertoz et al. }\cite{ertoz2004minds} present the MINDS framework for network intrusion detection and explain anomalies by association rules. Specifically, they first utilise an off-the-shelf anomaly detection model such as LOF \citep{breunig2000lof} to detect anomalous network connections. Second, they develop a Discriminating Association Pattern Generator to extract patterns that exclusively characterise normal instances or anomalous instances, respectively. The extracted patterns are human-comprehensible and thus serve as explanations for anomalies. Moreover, they attempt to assign anomalies to different groups based on the extracted patterns. {\color{black}Alternatively, Davidson }\cite{davidson2007anomaly} capitalizes on mixture modeling (a.k.a. model-based clustering) to perform anomaly detection. Specifically, for each data instance, this method calculates the likelihood of this instance belonging to each cluster. If the maximum obtained likelihood is less than a predefined threshold, the instance does not belong to any cluster and is therefore considered anomalous. Moreover, they describe a visualization approach to show normal and abnormal instances based on scatter plots, enabling end-users to quickly understand why an instance is considered anomalous. More importantly, they try to extract rules (in Conjunctive Normal Form) to describe each obtained cluster and those anomalies. By comparing these rules, one can easily understand why an anomaly is anomalous. 

Meanwhile, some model-specific  rule learners are proposed to extract a set of rules or patterns as the surrogate model, aiming to explain anomalies. For example, {\color{black}Das et al. }\cite{das2019active} define a novel formalism, known as \textit{compact description}, to extract rules to describe discovered anomalies. However, this method can only be applied to tree-based ensembles, and the extracted rules are represented using Disjunctive Normal Form. Moreover, they also develop an active anomaly explanation algorithm for generic ensembles, dubbed GLAD. For each detected anomaly, GLAD first identifies the base-learners (namely ensemble members) that contribute the most to the decision. Second, GLAD applies a model-agnostic explanation method such as LIME on these important base-learners, respectively, to generate explanations for the anomaly. Besides, {\color{black}Barbado et al. }\cite{barbado2022rule} apply several rule extraction techniques to OCSVM models \citep{scholkopf1999support} for anomaly explanation, and evaluate the quality of generated explanations accordingly. Specifically, these techniques first apply OCSVM to obtain normal and abnormal instances. Second, they use an off-the-shelf clustering method such as K-Prototypes \citep{ji2013improved} to iteratively divide the non-anomalous instances into different regions until no anomalies are contained in the generated regions. Third, since these regions are in the form of hypercubes, they can directly extract rules from the vertices of these hypercubes to explain why an instance is non-anomalous. More importantly, they define several metrics including comprehensibility, representativeness, stability and diversity to evaluate the quality of explanations. Besides, their methods can provide both local and global explanations. Although it is claimed that the whole process can be adapted to any anomaly detection model, this not shown.

Second, some rule learners are explored to extract decision trees as the surrogate model, aiming at explaining anomalies. For instance, {\color{black}Xu et al. }\cite{xu2009detecting} propose an approach to detect and explain system problems by mining console logs. Concretely, they leverage a Principle Component Analysis (PCA) based anomaly detection method \citep{dunia1997multi} to identify anomalies, followed by explaining the results using decision trees to mimic the decision-making process. However, {\color{black}Bin et al. }\cite{bin2016abnormal} show that using decision trees to explain the PCA model can be misleading, thereby failing to reveal the true decision-making process. Besides, {\color{black}Pevn{\`y} \& Kopp }\cite{pevny2014explaining} introduce a method called Explainer to explain anomalies using Disjunctive Normal Form (DNF). Specifically, given an anomaly, Explainer first trains a collection of trees known as Sapling Random Forests (SRF).  Each tree in an SRF is a binary decision tree with the aim of separating the anomaly from other normal instances. Second, once a tree is built, they utilise DNF to represent the path from the root node to the node that contains only the anomaly. Third, they aggregate the DNFs from all trees to a compact DNF to interpret the anomaly. Furthermore, {\color{black}Kopp et al. }\cite{kopp2014interpreting} extend Explainer by introducing two $k$-means based clustering methods to interpret anomalies when these anomalous instances form natural micro-clusters.

Third, some researchers attempt to utilise well-studied opaque models as surrogate models, and then leverage additional explanation techniques such as SHAP to explain surrogate models. For example, to monitor the average fuel consumption of fleet vehicles, {\color{black}Barbado }\cite{barbado2020anomaly} sets up an unsupervised anomaly detection process capable of explaining decisions through feature importance. First, they leverage a threshold-based model to detect anomalies. Second, they utilize two types of surrogate models to explain anomalies, including black-box anomaly detection models with a post-hoc local explanation (XGBoost \citep{chen2016xgboost} and LightGBM \citep{ke2017lightgbm} with LIME or SHAP),  and transparent anomaly detection models (ElasticNet \citep{zou2005regularization} and EBM \citep{nori2019interpretml}). Third, they evaluate these surrogate models in terms of predictive power and explanatory power. Particularly, their explanation method can also integrate domain knowledge given by business rules or counterfactual recommendations. Further, {\color{black}Kiefer \& Pesch }\cite{kiefer2021unsupervised} put forward an ensemble based anomaly detection model combined with model-agnostic explanation technique to identify and interpret anomalies in financial auditing data. Specifically, they construct an ensemble architecture to incorporate a wide range of unsupervised anomaly detection models, attempting to identify different types of anomalies. To interpret anomalies, they propose a four-step method: synthetic oversampling of anomalies, supervised model approximation (using SVM or XGBoost), LIME based local explanation, and explanation post-processing (visualisation or natural language description).

\textit{Discussion:} As can be seen, most methods in this category leverage rule learners to extract a set of rules or patterns to describe anomalies. Importantly, the resulting rules are often represented using a Disjunctive Normal Form or Conjunctive Normal Form. Consequently, it might be relatively easy to evaluate the quality of the resulting explanations, but this depends on many factors.

\subsection{Miscellaneous Methods}
%An introduction of other methods, describing why we have this subsection. (To shorten)

In addition to subspace based methods and surrogate methods, \textit{miscellaneous methods} such as visualisation and Shapley values are often used to obtain feature importance as explanations. Moreover, pattern comparison is commonly explored in sequential data to interpret anomalies in a post-hoc manner. Due to the diversity of these methods, we again organize them according to the type of data to which they are applicable.

%\subsubsection{Feature importance based methods for tabular data}
\subsubsection{Models for Tabular Data}

A wide range of methods has been proposed to explain anomalies in tabular data by showing feature contribution or selecting a subset of features. Importantly, some of these methods are model-agnostic. For instance, {\color{black}Liu et al. }\cite{liu2017contextual} introduce the COIN framework that consists of four main steps. For each anomaly, they first find its neighbours based on a distance measure such as Euclidean distance. Second, they leverage existing clustering algorithms to subdivide the anomaly and its neighbours into multiple disjoint clusters. Third, they apply a strategy such as synthetic sampling to expand the size of the anomaly cluster where the anomalous instance is located. Fourth, they train a simple classifier to separate these clusters, deriving an anomaly score and feature contributions from the parameters of the classifier. Moreover, COIN can also incorporate prior knowledge into the explanation process. {\color{black}Importantly, Siddiqui et al. }\cite{siddiqui2019sequential} present Sequential Feature Explanations (SFEs) to explain detected statistical outliers. Concretely, given an anomaly identified by any density-based detector, SFEs sequentially present a feature to the analyst until the analyst can confidently identify this anomaly. As a result, these features used to identify the anomaly constitute the corresponding explanations. Particularly, {\color{black}Siddiqui et al. }\cite{siddiqui2019detecting} apply Isolation Forest \citep{liu2008isolation} to detect cyber attacks and leverages SFEs to generate explanations. 

Shapley value based methods are often leveraged to obtain feature importance as explanations. For example,  {\color{black}Park et al. }\cite{park2020explainable} employ SHAP \citep{lundberg2017unified} to explain anomalies by showing feature contributions. Concretely, they set up an anomaly detection model by using random forest and employ the SHAP approach to explore the relationship between model results and input variables to generate explanations. Similarly, {\color{black}Kim et al. }\cite{kim2021explainable} utilize Isolation Forest on sensor stream data of marine engines to keep track of unusual engine conditions. Moreover, they leverage SHAP to identify which sensor is in charge of each abnormal data event and to quantify its contribution to the observed anomaly. Besides, using reconstruction errors as a measure to detect and explain anomalies is a common practice in unsupervised anomaly detection. However, {\color{black}Takeishi  }\cite{takeishi2019shapley} argues that by simply looking at the reconstruction error of each feature, one may fail to find the true cause of the anomaly. This is because a large reconstruction error in one feature may stem from another feature. To mitigate this problem, a method is introduced to compute the Shapley values \citep{sundararajan2020many} of reconstruction errors for PCA based anomaly detection method. The numerical examples show that the Shapley values are superior to reconstruction errors for explaining an anomaly. 

Model-specific techniques have also been developed to explain anomalies in tabular data, especially for Isolation Forest \citep{liu2008isolation}, a state-of-the-art anomaly detection model. For example, {\color{black}Kartha et al. }\cite{kartha2021you} develop a method to interpret anomalies identified by Isolation Forest by exploring the internal structure of an Isolation Forest to generate a feature importance vector, indicating the contribution of each feature to the anomaly score. Similarly, {\color{black}Carletti et al. }\cite{carletti2020interpretable2} propose DIFFI  to obtain feature importance scores for explaining Isolation Forest. Specifically, DIFFI provides a global feature importance score for each feature, indicating how that feature affects the overall decisions of Isolation Forest on the training data. Meanwhile, they present a local version of DIFFI, named local-DIFFI, to provide a local feature importance score for each feature, describing how each feature participates in making individual decisions on the test data. Importantly, they develop a feature selection method for unsupervised anomaly detection problems on this basis. Particularly, {\color{black}Carletti et al. }\cite{carletti2020interpretable} apply DIFFI on  real-world semiconductor manufacturing data to demonstrate its effectiveness. 

%\subsubsection{Pattern comparison and visualisation based methods for sequential data}

\subsubsection{Models for Sequential Data}

A common strategy to explain anomalies in sequential data is to contrast the observed pattern with its expected pattern or normal patterns. For instance, {\color{black}Babenko \& Pastore }\cite{babenko2009ava} leverage LFA \citep{mariani2008automated} to detect anomalies in system logs. On this basis, they present the so-called Automata Violation Analyzer (AVA) to  automatically explain anomalies detected by LFA. Specifically, AVA provides \textit{basic explanations} by comparing the expected event sequence with the observed event sequence, generating relatively simple explanations for the anomalous events. Furthermore, they combine these \textit{basic explanations} to obtain \textit{composite explanations}. Finally, they order these basic and composite explanations by their likelihood of explaining differences between the expected  event sequences and the observed event sequences. In addition, {\color{black}Leue \& Befrouei }\cite{leue2012counterexample} design a method to explain counterexamples that are symptoms of deadlocks in concurrent systems. These counterexamples can be considered as anomalies and the authors use sequential pattern mining to produce explanations for these anomalies. Specifically, they extract fixed-length common substrings from anomalous sequences and contrast them with normal sequences to explain the occurrence of anomalies. 

Meanwhile, leveraging visualization to explain anomalies in sequential data is also a common practice. For example, {\color{black}Rieck \& Laskov }\cite{rieck2009visualization} propose a technique for explaining intrusion detection results. Specifically, they present two methods for anomaly detection, viz. global anomaly detection and local anomaly detection. For each payload, the global anomaly detection method computes its distance to the center of all payloads as its anomaly score; In contrast, the local anomaly detection method computes the average distance to its k-nearest neighbors as its anomaly score. To explain an anomaly, they present a visualization tool to show the feature differences between the anomalous payload and the normal payloads. A large difference in a feature means that the corresponding feature value is anomalous. Furthermore, they also highlight the network content corresponding to the feature value in the original payload. Besides, {\color{black}Alizadeh et al. }\cite{alizadeh2021vehicle} implement an AutoRegressive Integrated Moving Average (ARIMA) based model together with a Virtual Reality (VR) tool to detect and interpret abnormal vehicle operating states. Specifically, modern vehicles are often equipped with multiple sensors to collect data used to monitor their operating status. To detect anomalies in such multi-channel time series data, they develop an ARIMA model for each channel (i.e., for each individual univariate time series). Hence, a large difference between the actual value and the predicted value indicates an anomaly. Importantly, they build a VR tool to visualize residuals from ARIMA models, aiming to better understand anomalies. Moreover, {\color{black}Markou et al. }\cite{markou2017use} create a tool for exploiting internet data to find abnormalities in transportation networks and connecting them to unique events. First, the baseline normality corresponding to GPS data for taxi journeys in New York City is trained based on historical mobility data. Next, they scan various days to look for days where demand deviates greatly from normality in order to identify abnormalities. To investigate the severity of daily traffic abnormalities, they consider the Z-score formula of kernel density values. The current traffic situation is considered abnormal if the Z-score value exceeds a given threshold. To explain the anomaly, they diagram the time and place of the anomaly and utilize that information to look for nearby unusual events using Google Searches.

\begin{landscape}
\small
\begin{longtable}{p{0.03\textwidth}p{0.02\textwidth}p{0.03\textwidth}p{0.23\textwidth}p{0.18\textwidth}p{0.03\textwidth}p{0.31\textwidth}p{0.32\textwidth}
}
\caption {Summary of surveyed shallow post-model techniques. \textit{Spec} indicates whether a method is model-agnostic (A) or model-specific (S). \textit{Pers} specifies whether a method is feature-based (F), sample-based (S) or pattern-based (P).  \textit{Tech} indicates the techniques used in each method. \textit{Data} represents the type of data to which each method can be applied. \textit{Data} indicates the data type for which the method is applicable (TN: Tabular Numeric; TC: Tabular Categorical; TM: Tabular Mixed; UTS: Univariate Time Series; MTS: Multivariate Time Series; ES: Event Sequence). \textit{Loc} shows whether a method provides a local explanation (L) or global explanation (G). \textit{Pros} and \textit{Cons} describe the advantages and disadvantages of each method, respectively.}
\label{tab:summary_shallow_post}\\
\hline Ref & Spec & Pers & Tech & Data &Loc  & Pros & Cons\\ 
\hline
\cite{knorr1999finding} & A & F & Others (Subspace)  & Static tabular  & L & Applicable to any AD;  Good explanation fidelity (exception in post-hoc)  & High computational cost  \\
\cite{zhang2004hos} & A & F & Others (Subspace) & Static tabular   & L & Applicable to any AD & Poor explanation fidelity; Compares scores in subspaces with different dimensionalities\\
\cite{angiulli2009detecting} & A & F & Others (Subspace) & Static TC  &  L \& G & Applicable to any AD & Poor explanation fidelity; Only applicable to categorical data\\
\cite{muller2012outrules} & A & F & Others (Subspace + Context) & Static TN    & L & Applicable to any AD; Considers multiple contexts & Poor explanation fidelity \\
\cite{micenkova2013explaining} & A & F & Others (Subspace) & Static TN  & L & Applicable to any AD & Poor explanation fidelity \\

\cite{duan2015mining} & A & F & Others (Subspace) & Static TN  & L & Applicable to any AD & Poor explanation fidelity; KDE does not work with high-dimensional data\\
\cite{kuo2016framework} & A & F & Others (Subspace) & Static TN/TC  & L & Applicable to any AD; Enables human in the loop & Poor explanation fidelity; Poor scalability\\
\cite{vinh2016discovering} & A & F & Others (Subspace) & Static TN   & L& Applicable to any AD; Fast search; Dimensionality unbiased  & Poor scalability in high-dimensional data; Poor explanation fidelity\\
\cite{gupta2018beyond} & A & F & Visualisation  & Static TN   & L & Applicable to any AD; Easy to understand by non-experts & Only considers 2D subspace\\
\cite{angiulli2017outlying} & A & F  & Others (Subspace + Contextual Contrast) & Static TM   & L & Applicable to any AD; Considers both numeric and categorical features & KDE does not work well in high-dimensional subspace\\
\cite{macha2018explaining} & A & F & Others (Subspace + Rule Extraction) & Static TN  & L & Applicable to any AD; Explains anomalies in groups & Poor explanation fidelity\\
\cite{ertoz2004minds} & A & P & Approximate (Rule Extraction/Association Rule Mining)  & Static ES  & L & Applicable to any AD & Poor explanation fidelity\\
\cite{davidson2007anomaly} & S & S & Approximate (Rule Extraction) + Visualisation  & Static TM   & G & Provides visual explanations & Only applicable to clustering based AD\\
\cite{xu2009detecting} & A & F  & Approximate (Rule Extraction/Decision Trees) + Visualisation & Static logs  & G & Good scalability & Post-hoc explanations can be misleading\\
\cite{pevny2014explaining} & A & F &  Approximate (Rule Extraction/Decision Trees)  & Static TN  & L & Applicable to any AD; Good scalability & Poor explanation fidelity \\
\cite{das2019active} & S  & F & Approximate (Rule Extraction) & Static and streaming TM  & L & Able to handle streaming data & Only applicable to tree-based ensembles\\
\cite{barbado2022rule} & A & F & Approximate (Rule Extraction) & Static TM  & L \& G & Easy to evaluate the explanation quality  & Only considers OCSVM \\
\cite{barbado2020anomaly} & A & F & Approximate ( XGBoost/LightGBM + LIME/SHAP; ElasticNet/EBM)   & Static MTS  & G & Able to integrate domain knowledge & Not suitable for explaining only one anomalous point \\
\cite{kiefer2021unsupervised} & A & F & Approximate (SVM/XGBoost + LIME) + Visualisation  & Static UTS  & G & Applicable to any AD; End-user dependent explanations  &  Poor explanation fidelity \\
\cite{babenko2009ava} & S & P & Others (Pattern comparison) & Streaming ES  & L & Able to handle streaming data & Only applicable to LFA-like AD \\
\cite{leue2012counterexample} & A & P & Others (Pattern comparison)  & Static ES & L & Generalisable & Needs many anomalies\\
\cite{rieck2009visualization} & S  & F & Visualization  & Static ES  & L & Provides visual explanations & Only applicable to distance based AD methods \\
\cite{liu2017contextual} & A & F & Others (Neighbours + Data Augmentation + Classification + Feature Contribution)  & Static TN   & L & Incorporates prior knowledge; Applicable to various anomaly detectors & Poor explanation fidelity; Important parameters to set by user; Only considers individual anomalies\\
\cite{siddiqui2019sequential} & S & F & Others (Separation-Based)  & Static TN  & L & Provides quantitative evaluation of explanation quality  & Only applicable to density-based AD\\
\cite{kartha2021you} & S & F & Others (Isolation based feature importance) & Static TM & L & High explanation fidelity & Only applicable to Isolation Forest \\
\cite{carletti2020interpretable2} & S & F & Others (Isolation based feature importance)  &  Static TM  & L \& G & Provides local and global explanations; High explanation fidelity & Only applicable to Isolation Forest \\
\cite{alizadeh2021vehicle} & S & P & Visualisation  & Static MTS   & L & Interpretable statistical model and visual interpretation & Only applicable to ARIMA\\
\cite{markou2017use} & S & F & Visualization  & Static TS  & L & Provides visual explanations & Only applicable to Spatiotemporal model \\
\cite{takeishi2019shapley} & S & F & Shapley values of reconstruction errors & Static tabular   & L & More reliable compared to reconstruction error based explanation & Only applicable to PCA \\
\cite{park2020explainable} & A & F & SHAP  & Static tabular  & L & Applicable to any AD &  Poor explanation fidelity\\
\cite{kim2021explainable} & A & F & SHAP  &  Streaming MTS & L & Applicable to any AD & Poor explanation fidelity\\

\hline  
\end{longtable}
\end{landscape}

\textit{Discussion:} This subsection reviewed a wide range of shallow XAD techniques that are explored to interpret anomalies in a post-hoc manner. In contrast to \textit{subspace based methods} and \textit{surrogate methods}, the techniques investigated here vary by data type. For instance, Pattern Comparison and Visualisation are commonly utilized to explain anomaly in sequential data. Methods in this group usually do not have an explicit \textit{Explanation-Definition} of anomalies since they directly illustrate the anomalies by comparing patterns or using visualisation tools. Meanwhile, feature importance that is obtained by using SHAP techniques, separation or isolation-based measure, plays an important role in explaining anomalies in tabular data. Using Shapley values techniques such as SHAP to obtain feature importance usually does not require a definition of anomaly. Moreover, model-specific techniques such as References \citep{kartha2021you,carletti2020interpretable2,alizadeh2021vehicle} generally have consistent \textit{Explanation-Definition} and \textit{Detection-Definition} of anomaly as they explore the internal {\color{black} structure of an anomaly detection model} to generate explanations. On the contrary, model-agnostic techniques such as Reference \citep{liu2017contextual} usually have an \textit{Explanation-Definition} that may differ from the \textit{Detection-Definition}.

\subsection{Summary}
While Table \ref{tab:summary_shallow_post} provides the full characterization for all methods discussed in this section based on the six criteria of our taxonomy, we here make some general observations on shallow post-model XAD techniques. 

First, nearly all \textit{subspace based methods} and \textit{surrogate methods} are model-agnostic in the sense that they are applicable to any anomaly detection model or given anomalies. In other words, these methods do not explore the internal structure of an anomaly detection model and therefore cannot have a full grasp of the underlying decision-making mechanism, rendering the provided explanations less useful and potentially resulting in weak interpretability. In contrast, \textit{miscellaneous methods} are mainly model-specific, as they explore the internal structure of anomaly detection models to generate feature importance as explanations. As a result, the obtained explanations are more reliable and actionable.

Second, all these methods provide feature-based explanations (including pattern-based explanations) except that {\color{black}Davidson }\cite{davidson2007anomaly} provides sample-based explanations. We consider the lack of sample-based explanation methods to be a gap in the literature that might be of interest for future research.

Third, most shallow post-model XAD techniques, especially \textit{subspace based methods} and \textit{miscellaneous methods}, can only provide local explanations. In other words, they can merely interpret an individual anomaly at a time. As a result, the explanation may be highly sensitive to noise or biased since the employed XAD methods are short of a holistic perspective on the decision-making process and logic.

\section{Literature Review on Deep Post-Model Techniques}
%[Illustrating the advantages of using DNN to perform anomaly detection; emphasising the shortcomings of using DNN to conduct anomaly detection, namely poor interpretability; enumerating commonly used DNN anomaly detection methods, including AE, LSTM, CNN, etc.; discussing popular XAI techniques used in DNN]

Deep learning, based on artificial neural networks, has become prevalent in anomaly detection due to its capability to learn expressive feature representations and/or anomaly scores for complex data such as text, audio, images, videos and graph \citep{pang2021deep}. A wealth of deep anomaly detection methods, including those based on AutoEncoders (AE), Long Short-Term Memory (LSTM), Convolutional Neural Network (CNN), Generative Adversarial Network (GAN) and other neural networks, have been proposed and have been shown to be more accurate than traditional methods when it comes to detecting anomalies in complex data. However, although deep anomaly detection methods tend to have high detection accuracy, they are often criticized for their poor interpretability. For this reason, some studies have attempted to leverage post-hoc XAI techniques to improve the interpretability of corresponding neural networks. Importantly, which XAI techniques are available may vary depending on the specific neural network used. For instance, AE based models typically employ reconstruction errors to explain anomalies, while LSTM based models generally leverage SHAP techniques to interpret anomalies. Therefore, we will present the review results according to the type of neural network used to perform anomaly detection, which is correlated with the data type that can be used (e.g., CNNs for images, RNNs for sequential data).

\subsection{Explaining AutoEncoders}
%[Introduction of AE] 

An AutoEncoder (AE) is a type of neural network that first encodes the given data instances into some low-dimensional feature representation space and then decodes them back under the constraint of minimizing the reconstruction error. Several types of AEs have been introduced, including vanilla AE such as replicator neural network, Sparse AutoEncoders (SAE), Denoising AutoEncoders (DAE), Contractive AutoEncoders (CAE), Variational AutoEncoders (VAE), and other variants \citep{bank2020autoencoders}. AEs are widely used for anomaly detection, based on the assumption that anomalies are more difficult to reconstruct from the compressed feature representation space than normal instances. 

First of all, Shapley values based techniques such as SHAP are typically used to obtain feature contributions for explaining AEs. For instance,  {\color{black}Giurgiu \& Schumann }\cite{giurgiu2019additive} extend SHAP to explain anomalies identified via a GRU-based AutoEncoder in multivariate time series data. Specifically, they modify kernel SHAP \citep{lundberg2017unified} to output the windows that contribute the most to the anomaly and also the windows that counteract the most to the anomaly as explanations. Besides, to detect and explain anomalies in mobile Radio Access Network (RAN) data,  {\color{black}Chawla et al. }\cite{chawla2020interpretable} set up a Sparse AutoEncoder (SAE) based anomaly detection algorithm and then applies kernel SHAP to explain the results. Furthermore, {\color{black}Jakubowski et al. }\cite{jakubowski2021anomaly} propose a Variational AutoEncoder (VAE) model combined with Shapley values to detect and interpret anomalies in an asset degradation process. Concretely, they compute Shapley values to generate both local and global explanations for anomalies. {\color{black}Additionaly, Serradilla et al. }\cite{serradilla2021adaptable} utilise different machine learning approaches to detect, predict and explain anomalies in press machines to achieve predictable maintenance. To interpret an anomaly detected by AutoEncoder (AE), they first leverage t-SNE \citep{van2008visualizing} to visualise the learned latent feature spaces. Next, they employ the GradientExplainer tool \citep{lundberg2021game}, which combines SHAP, Integrated Gradients \citep{sundararajan2017axiomatic}, and SmoothGrad \citep{smilkov2017smoothgrad}, to analyze which input features are associated with the anomaly. 

Second, many methods attempt to track reconstruction errors to obtain feature contribution by exploring the internal structure of AEs. Therefore, these methods are generally model-specific. For instance, {\color{black}Ikeda et al. }\cite{ikeda2018anomaly} design a  Multimodal AutoEncoder (MAE) model to detect anomalies emerging in ICT systems. More importantly, by using sparse optimization, they also propose an algorithm to estimate the contributing dimensions in an AE to anomalies as explanations. Besides, {\color{black}Nguyen et al. }\cite{nguyen2019gee} introduce a framework called GEE to detect and explain anomalies in network traffic. Specifically,  they train a Variational AutoEncoder (VAE) model on a normal dataset  to learn the normal behaviour of a network, and then employ gradient-based fingerprinting technique to identify the main features causing the anomaly. Similarly, {\color{black}Memarzadeh et al. }\cite{memarzadeh2022multiclass} propose a deep generative model based on VAE. Particularly, they achieve model interpretability by evaluating feature importance through the random-permutation method. Additionally, {\color{black}Chen et al. }\cite{chen2021daemon} put forward DAEMON, which trains an Adversarial AutoEncoder (AAE) to learn the typical pattern of multivariate time series, and then use the reconstruction error to identify and explain anomalies. Meanwhile, to monitor wireless spectrum and identify unexpected behavior, {\color{black}Rajendran et al. }\cite{rajendran2018saife} present an AAE based anomaly detection method named SAIFE. Since the AAE is trained in three phases, viz. \textit{reconstruction}, \textit{regularization} and \textit{semi-supervised} \citep{makhzani2015adversarial}, SAIFE attempts to localize the anomalous regions based on the reconstruction errors coupled with the semi-supervised features, providing explanations for the anomalies. Furthermore, {\color{black}Ikeda et al. }\cite{ikeda2018estimation} set up an anomaly detection model based on VAE, and then estimate the features that contribute the most to the identified anomalies as explanations. Concretely, they present an approximative probabilistic model based on the trained VAE to estimate contributing features via exploring the so-called \textit{true latent distribution}. The \textit{true latent distribution} defines how an anomalous instance would be if it were normal. Importantly, they argue that directly estimating feature contribution based on the deviating latent distribution or reconstruction errors will lead to high false positives and/or negatives.

Third, some researchers attempt to utilise surrogate models such as LIME and rule learners to explain AEs. For example, {\color{black}Wu \& Wang }\cite{wu2021locally} propose a neural network based model incorporating LIME techniques to detect and interpret fraudulent credit card transactions. Specifically, the anomaly detection model contains an AE and a Multilayer Perceptron (MLP) classifier, which are trained in an adversarial manner. To interpret an anomaly, they apply three independent LIME based models to explain the AE, MLP and AE \& MLP models, respectively. Besides, {\color{black}Song et al. }\cite{8637456} develop the EXAD system to identify and interpret anomalies from Apache Spark traces. First, the EXAD system adapts AE and LSTM to perform anomaly detection. Second, they propose three ways to explain anomalies. The first one is to build a conjunction of the atomic predicates, which can be solved by a greedy algorithm but cannot guarantee the performance. To overcome this limitation, the second one attempts to use an entropy based reward function to build atomic predicates. Furthermore, they present these constructed predicates in a Conjunctive Normal Form. The third one is to approximate the anomaly detection neural networks using a decision tree. From the decision tree, they generate explanations in a Disjunctive Normal Form. Additionally, {\color{black}De Moura et al. }\cite{de2021interpretable} present the Lane Change Detector (LCD) model to detect and explain when the surrounding vehicles of an ego vehicle change their lanes.  Specifically, the LCD model consists of three independent AE models trained on three different datasets. On this basis, they set up a decision rule set based model by extracting rules from the reconstruction errors produced by these three separate models, to determine when an anomaly happens. Besides, {\color{black}Gnoss et al. }\cite{gnoss2022xai}  first annotate journal entries with previously trained AutoEncoders, and then train three XAI models using these annotations. First, they utilise Decision Tree and Linear Regression, two intrinsically interpretable models, to simulate AE. The feature importance values of Decision Tree and the odd ratio values are calculated to show which feature is relevant to the anomalies. Additionally, they also leverage SHAP to explain the AE model.

Fourth, visualisation techniques such as Heatmaps and Saliency Maps are often constructed to help explain AEs. For instance, {\color{black}Kitamura \& Nonaka }\cite{kitamura2019explainable} set up an encoder-decoder based model to detect anomalies in images. To generate explanations for an anomaly,  they first develop a feature extractor that is trained on a dataset consisting of normal images and their corresponding reconstructed images. Second, using this feature extractor to extract latent features, their method attempts to find the difference on the feature-level between the input image and the reconstructed image. On this basis, their method localizes and visualizes abnormal regions as explanations for the anomaly. Besides, {\color{black}Feng et al. }\cite{feng2021anomaly} develop a Two-Stream AutoEncoder (AE) based model to detect abnormal events in videos and then utilise a Feature Map Visualization method to interpret the anomalies. Moreover, {\color{black}Guo et al. } \cite{guo2021interpretable} set up a Sequence-to-Sequence VAE based model to detect anomalies in event sequences. To reveal anomalous events, they investigate the differences between the anomalous sequence together with its reconstructed sequence and a set of normal sequences close to the anomalous sequence in the latent space. Importantly, they build a visualization tool to facilitate the comparisons. In addition, {\color{black}Szymanowicz et al. }\cite{szymanowicz2022discrete} develop a method for detecting and automatically explaining anomalous events in video. They first design an encoder-decoder architecture based on U-Net \citep{ronneberger2015u} to detect anomalies, thereby generating saliency maps by computing per-pixel differences between actual and predicted frames. Second, based on the per-pixel squared errors in the saliency maps, they introduce an explanation module that can provide spatial location and human-understandable representation for the identified anomalous event.

Finally, a wide range of methods such as feature selection, Markov Chain Monte Carlo, and providing similar historic anomalies, are also explored to facilitate the interpretability of AE based anomaly detection. For example, {\color{black}Chakraborttii \& Litz }\cite{chakraborttii2020explaining} develop an AE based model to detect Solid-State Drive (SSD) failures. To produce explanations, they investigate the reconstruction error per feature, wherein a feature with a reconstruction error greater than the average error is considered a significant cause. Particularly, they apply three types of feature selection techniques, viz. Filter, Wrapper and Embedded, to select important features to train the AE model, facilitating the interpretability of resulted anomaly detection model. Besides, {\color{black}Li et al. }\cite{li2021vaga} develop a Variational AutoEncoder (VAE) and genetic algorithm (GA) based framework, called VAGA, to detect anomalies in high-dimensional data and search corresponding abnormal subspaces. Concretely, for each identified anomaly, they utilize a GA to search the subspace where the anomaly deviates most. Additionally, {\color{black}Li et al. } \cite{li2021multivariate} introduce \textit{InterFusion}, a model based on hierarchical Variational AutoEncoder (HVAE) and Markov Chain Monte Carlo (MCMC) for detecting and explaining anomalies in multivariate time series data. Specifically, given an anomaly, they set up a MCMC-based method to find a set of the most anomalous metrics as explanations. Furthermore, {\color{black}Assaf et al. }\cite{assaf2021anomaly} develop a Convolutional AutoEncoders (ConvAE) based anomaly detection method and an explainability framework to detect and explain anomalies in data storage systems, respectively. Particularly, for each anomaly, they attempt to use cosine similarity over the embedding space to find similar historical anomalies, thereby explaining the anomaly through association. 

\textit{Discussion:} AEs are the most widely used deep learning method to detect anomalies in tabular data, sequence data, image data, video data and graph data. As a result, a plethora of methods are also proposed to explain AEs. Concretely, XAD techniques such as reconstruction error-based feature contribution, Kernel SHAP, GradientExplainer, LIME, rule extraction and feature map visualisation are often leveraged to obtain explanations. Importantly, most of these explanation methods only provide weak interpretability, as they only explain a single anomaly at a time by exploring some important properties of AE-based detection models.

\subsection{Explaining Recurrent Neural Networks}
%[Introduction of RNN]
A Recurrent Neural Network (RNN) is a specific type of neural network that is capable of learning features and long term dependencies in sequential data \citep{salehinejad2017recent}. Specifically, sequential data refers to any data that is ordered into sequences, including time series, text streams, DNA sequences, audio clips, video clips, etc. To address the different challenges of modeling sequential data, various RNN architectures have been proposed. More concretely, frequently used RNNs include deep RNNs with Multi-Layer Perceptron, Bidirectional RNN (BiRNN), Recurrent Convolutional Neural Networks (RCNN), Multi-Dimensional Recurrent Neural Networks (MDRNN), Long-Short Term Memory (LSTM), Gated Recurrent Unit (GRU), Memory Networks, Structurally Constrained Recurrent Neural Network (SCRNN), Unitary Recurrent Neural Networks (Unitary RNN), etc. Particularly, by assuming normal instances are temporally more predictable than anomalous instances, RNNs are extensively used to identify anomalies in sequential data because of their ability to model temporal dependencies.

First of all, Shapley values based techniques such as SHAP are the most typical method used to obtain feature contributions, aiming to explain anomalies identified by RNNs. For instance, {\color{black}Tall{\'o}n-Ballesteros \& Chen }\cite{tallon2020explainable} utilise Decision Trees \citep{chen2004failure} and DeepLog \citep{du2017deeplog} to detect anomalies in system logs, and then explain the results using the Shapley value approach. To explain an anomaly, they treat each event in the logs as a player without examining the model structure to generate Shapley values. Moreover, {\color{black} Hwang \& Lee }\cite{hwang2021sfd} propose a bidirectional stackable LSTM-based  anomaly detection model for industrial control system anomaly detection. For each identified anomaly, they employ SHAP values to obtain a contribution score of each feature as an explanation. Similarly, {\color{black} Jakubowski et al. }\cite{jakubowski2021explainable} examine the issue of anomaly detection when hot rolling slabs into coils. They utilise LSTM to construct a modified AutoEncoder architecture in order to find anomalies. Importantly, they are able to pinpoint the origin of the majority of the abnormalities identified by the deep learning model through analysis of the SHAP interpretation. Furthermore, {\color{black} Nor et al. }\cite{nor2021application} present a probabilistic LSTM based model combined with SHAP to detect and interpret anomalies in gas turbines. More importantly, they evaluate the quality of post-hoc explanations from two aspects, viz. \textit{local accuracy} and \textit{consistency}. Specifically, \textit{local accuracy} describes the relationship between feature contributions and predictions, while \textit{consistency} checks whether the interpretation is consistent with changes in the input features. 

Second, some researchers attempt to utilise surrogate models such as LIME to explain RNN. For example, {\color{black}Herskind Sejr et al. }\cite{herskind2021outlier} create a predictive neural network-based unsupervised system by training an LSTM model and use reconstruction errors to assess data abnormalities. Importantly, the system offers two layers of anomaly interpretation: deviations from model predictions, and interpretations of model predictions, in order to make the process transparent to developers and users. They employ Mean Absolute Error to illustrate how observations diverge from assumptions at the first level. For the second level, they simulate a black-box model to provide an explanation using LIME. Additionally, {\color{black}Mathonsi \& van Zyl } \cite{mathonsi2021statistics} present Multivariate Exponential Smoothing Long Short-Term Memory (MES-LSTM) that combines statistics and deep learning. Particularly, they integrate SHAP and LIME and introduce a metric---called Mean Discovery Score---that aims to show which predictors are most strongly associated with the anomalies. 

Third, other methods such as Layer-wise Relevance Propagation (LRP), Integrated Gradients, and Attention Mechanism, are also leveraged to explain RNN based anomaly detection. For instance, due to the complexity of log systems and the unstructured nature of the resulting logs, {\color{black}Patil et al. }\cite{patil2019explainable} use LSTM to detect anomalies in such systems. To generate explanations for each identified anomaly, they utilise LRP to generate relevance scores for every feature at every timestep. Moreover, {\color{black}Han et al. }\cite{han2021interpretablesad} present \textit{InterpretableSAD}, a Negative Sampling based method for detecting and interpreting anomalies in sequential log data. First, due to the scarcity of anomalous instances, they adapt a data augmentation strategy via negative sampling to generate a dataset that contains sufficient anomalous samples. Second, they train a LSTM model based on this augmented labelled dataset. Third, they apply Integrated Gradients to identify anomalous events that lead to the outlyingness. Furthermore, to detect anomalies in system logs, {\color{black}Brown et al. }\cite{brown2018recurrent} implement four attention mechanisms in LSTM and prove that compared to Bidirectional LSTM, the attention mechanism augmented LSTM not only retains high performance, but also provides information about feature importance and relationship mapping between features, which provides explainability.

\textit{Discussion:} RNNs are primarily employed to detect anomalies in sequence data. Typical XAD techniques for interpreting anomalies identified by RNN-based models include Shapley-value-based methods, surrogate models, and other versatile techniques such as Layer-wise Relevance Propagation, Integrated Gradients, and Attention Mechanism. These post-hoc explanation methods are usually computationally expensive, making it difficult to provide real-time explanations.

%comment: The Discussion paragraphs in Section 7 are significantly shorter than those in earlier sections. It would be to extend them somewhat, to give more insights and make them more balanced throughout the manuscript.

\subsection{Explaining Convolutional Neural Networks}
%[Introduction of CNN]
A Convolutional Neural Network (CNN) is a specific type of neural network inspired by the visual cortex of animals. CNNs are widely used in computer vision field because of its strong ability to extract  features from image data with convolution structures. Moreover, CNNs have also been shown to be useful for extracting complex hidden features in sequential data \citep{gorokhov2017convolutional}. Accordingly, a variety of CNN architectures have been proposed, including LeNet, AlexNet, GoogleNet, VGGNet, Inception V4, ResNet, etc. Some studies have attempted to utilize CNNs for anomaly detection, especially in the fields of intrusion detection, image anomaly detection, etc.

First, one line of research attempts to utilise surrogate models such as LIME and rule learners to explain CNN. For example, {\color{black}Cheong et al. }\cite{cheong2021interpretable} set up a SpatioTemporal Convolutional Neural Network–based Relational Network (STCNN-RN) to detect anomalous events in financial markets. For each anomaly, they apply LIME to provide a local explanation by indicating the contribution of each feature. Besides, {\color{black}Levy et al. }\cite{levy2022anomili} propose an end-to-end anomaly detection model named AnoMili, which can also provide real-time explanations. Specifically, AnoMili consists of four stages. First, they introduce a physical intrusion detection mechanism by using AutoEncoder (AE). Second, if no anomalous device is discovered, they train a CNN-based classifier on voltage signals of each device, aiming to detect spoofing attacks. Third, they utilise LSTM to build a context-based anomaly detection mechanism, which detects anomalous messages based on their context. Finally, to interpret an anomalous message, they leverage decision tree to locally approximate the detection result and also apply SHAP TreeExplainer \citep{lundberg2017unified} to identify the most important features in real-time.

Second, visualisation techniques are often combined with other techniques such as Gradient Backpropagation and Layer-wise Relevance Propagation to explain CNN based anomaly detection. For instance, {\color{black}Saeki et al. }\cite{saeki2019visual} present a CNN based method to detect and explain machinery faults based on vibration data. For each detected anomaly, they utilize grad-CAM \citep{selvaraju2017grad}, which is a gradient-based localization approach, to obtain an importance map in the feature space. Fourth, they combine the results of grad-CAM with a visualization approach called Guided Backpropagation \citep{springenberg2014striving}. Concretely, this method can visualize the predictions via backpropagation from the output space to the input space, generating explanations for the anomaly. Moreover, {\color{black}Chong et al. }\cite{chong2021toward} introduce a CNN based Teacher–Student Network based model combined with Layer-wise Relevance Propagation (LRP) technique to detect and explain anomalies. To interpret an anomaly, they provide an example-based explanation by showing its top prototypes (namely top nearest neighbours). Importantly, they apply LRP to show a pixel-level similarity between the anomaly and each of its top prototypes. Additionally, {\color{black}Szymanowicz et al. }\cite{szymanowicz2021x} introduce a model to detect and explain anomalies in videos. Specifically, they implement R-CNN to detect objects in video, and then employ Dual Relation Graph for human-object interaction recognition. The video is encoded with a collection of human-object interaction vectors (HOI vectors) for each frame. When the likelihood of the HOI vector in a scenario is less than a threshold, an anomaly is proclaimed. After using PCA to reduce the dimension of non-anomalies, they train a Gaussian Mixture Model (GMM). A video frame is deemed abnormal if any of its HOI vectors are lower than the threshold probability under the GMM. The distance between the anomalous HOI vector and the usual HOI vector is then weighted and visualized as a 2D heatmap to help understand abnormalities.

Third, some researchers attempt to directly utilise the semantic anomaly scores as explanations. For instance, {\color{black}Hinami et al. }\cite{hinami2017joint} utilise a general CNN model and context-sensitive anomaly detectors to identify and explain abnormal events in films. Specifically, they set up a Fast R-CNN based model to learn multiple concepts in videos and then  extract semantic features. On this basis, they apply a context-sensitive anomaly detector to obtain semantic anomaly scores, which can be seen as explanations for anomalies. 

\textit{Discussion:} CNN-based anomaly detection models are mainly leveraged to detect anomalies in image data. To explain anomalies identified by CNNs, XAD techniques such as surrogate models (LIME and rule learners), Gradient Backpropagation, Layer-wise Relevance Propagation and visualisations are commonly used. However, some post-hoc explanation methods, especially surrogate models, may suffer from poor explanation fidelity. In other words, the generated explanations may not reflect the actual anomaly detection process of CNNs.

\subsection{Explaining other deep neural networks}
%[Introduction of other popular DNN in anomaly detection]

In addition to AEs, RNNs and CNNs, other deep neural networks---such as Generative Adversarial Networks (GANs), Deep OCSVM, and Deviation Network (DevNet)---can also be used for anomaly detection. Therefore, the interpretation of these types of networks is also relevant. 

First, some studies propose explanation methods for general DNNs. For instance, {\color{black}Amarasinghe et al. }\cite{amarasinghe2018toward} propose a framework for explainable Deep Neural Network (DNN) based anomaly detection. Specifically, they assume the anomaly detection is performed in a supervised setting and leverage LRP to obtain the input feature relevance for making a decision. %Besides, Reference \cite{sipple2020interpretable} proposes a negative sampling based approach to detect and explain device failures in the Internet of Things. For an anomaly detected by a Neural Network based model, they leverage Integrated Gradients techniques to attribute the anomaly score to each feature as explanations. 
Besides, {\color{black}Sipple }\cite{sipple2020interpretable} trains an anomaly detector using Neural Network with negative sampling to detect device failures in the Internet of Things. For each identified anomaly, they leverage Integrated Gradients techniques to attribute the anomaly score to each feature and provide a contrastive nearest normal instance as explanations.

Second, some researchers utilise techniques such as self-attention learning based feature selection or gradient back propagation based feature contribution to explain a Deviation Network. For instance, {\color{black}Xu et al. }\cite{xu2021beyond} propose Attention-guided Triplet deviation network for Outlier interpretatioN (ATON) to explain anomalies in a post-hoc fashion. Specifically, ATON is composed of two main modules, viz. the feature embedding module and the customized self-attention learning module. The feature embedding module transforms the original feature space into an embedding space with extended high-level information. Meanwhile, given an anomaly, the customized self-attention learning module can obtain the contribution of each learned feature to its separability. Based on the embedding module and the corresponding attention coefficients, they distill a subset of the original features that lead to the separability of the anomalous instance. Meanwhile, {\color{black}Pang et al. }\cite{pang2021explainable} put forward FASD, a weakly-supervised framework to detect anomalies when a few labeled anomalies of interest are available. Specifically, they instantiate this framework as a deviation networks (DevNet) model, which assumes that the anomaly scores of normal instances are drawn from a Gaussian prior distribution and the anomaly scores of anomalies come from the upper tail of the prior. To interpret an anomaly, they compute the contribution of each input feature to the final anomaly score through gradient-based back propagation. 

Third, deep Taylor decomposition \citep{montavon2017explaining} is leveraged to explain models such as OCSVM, KDE, etc. For example, {\color{black}Kauffmann et al. }\cite{kauffmann2020towards} first convert the OCSVM models to neural networks, and then they modify the deep Taylor decomposition method to be applicable to these neural networks. In addition, they show its superiority to other explanation methods such as Distance Decomposition, Gradient Based Method, SHAP Values and Edge Detection, which are commonly used in deep learning to produce pixel-wise explanations of decisions. However, this method itself has many parameters to tune when applied to different methods or datasets, sometimes rendering the explanation method itself not explainable. Moreover, it also makes many strong assumptions and approximations. Similarly, {\color{black}Kauffmann et al. }\cite{kauffmann2020clever} reveal the widespread occurrence of Clever Hans phenomena in unsupervised anomaly detection models. Concretely, they propose an XAI procedure based on Deep Taylor Decomposition to highlight relevant features for detecting anomalies, and apply it on models including AutoEncoder reconstruction based detectors, Deep One-Class and KDE based detectors, generating pixel-wise explanations of outlyingness.

\begin{landscape}
\small
\begin{longtable}
{p{0.03\textwidth}p{0.11\textwidth}p{0.05\textwidth}p{0.27\textwidth}p{0.12\textwidth}p{0.03\textwidth}p{0.25\textwidth}p{0.27\textwidth}
}
\caption {Summary of surveyed deep post-model techniques. \textit{Spec} indicates whether a method is model-agnostic (A) or model-specific (S). If it is model-specific, we also indicate the models to which it applies. However, for completeness, we also indicate the involved DNN framework for model-agnostic techniques. \textit{Pers} specifies whether a method is feature-based (F), sample-based (S) or pattern-based (P). \textit{Data} indicates the data type for which the method is applicable (TN: Tabular Numeric; TC: Tabular Categorical; TM: Tabular Mixed; UTS: Univariate Time Series; MTS: Multivariate Time Series; ES: Event Sequence). \textit{Loc} shows whether a method provides a local explanation (L) or global explanation (G). }
\label{tab:summary_deep_post}\\
\hline Ref & Spec & Pers & Tech & Data &Loc  & Pros & Cons\\
\hline
\cite{ikeda2018anomaly} & S (MAE) & F & Reconstruction error-based feature contribution using sparse optimization  & Static network-flow data  & L & Able to handle cross-domain data & Only applicable to AEs\\
\cite{giurgiu2019additive} & A (GRU-based AE) & S \& P &  Kernel SHAP based feature importance & Static MTS & L & Applicable to any AD & Assumes feature independence in Kernel SHAP\\
\cite{chawla2020interpretable} & A (SAE) & F & Kernel SHAP based feature importance & Streaming ES & L & Applicable to any AD & Assumes feature independence in Kernel SHAP\\
\cite{kitamura2019explainable} & S (AEs) & F & Feature-level reconstruction error + Visualisation & Static image & L & Provides visual explanations & Only applicable to AEs\\
\cite{chakraborttii2020explaining} & S (AEs) & F & Reconstruction error based feature contribution + Feature selection & Static telemetry logs & L & Able to handle evolving data & Only applicable to AEs \\
\cite{serradilla2021adaptable} & S (AEs) & P & GradientExplainer + Visualization & Static MTS  & L & Provides visual explanations & Only applicable to AEs \\
\cite{feng2021anomaly} & S (CNN-based AE) & F & Feature map visualization & Static video & L & Provides visual explanations & Post-hoc explanations may be misleading\\
\cite{wu2021locally} & A (AE+MLP) & F & Approximate (LIME) & Static TN & L & Applicable to any AD & Poor explanation fidelity \\
\cite{8637456} & A (AE+LSTM) & F & Approximate (Rule extraction) &  Streaming ES & L & Handles streaming data; Applicable to any AD & Post-hoc explanations may not be reliable \\
\cite{nguyen2019gee} & S (VAE) & F & Gradient based feature contribution  & Streaming ES  & L &  Handles streaming data & Only applicable to reconstruction based DNN\\
\cite{guo2021interpretable} & S (VAE)  & P &  Reconstruction error based pattern comparison + Visualization  & Static ES & L & Provides interactive visual explanations  & Only applicable to reconstruction based AD\\
\cite{li2021vaga} & A (VAE) & F & GA based subspace search &  Static TM  & L & Applicable to any AD & High computational cost\\
\cite{jakubowski2021anomaly} & A (VAE) & F & Shapley values based feature contribution & Static MTS & L \& G & Applicable to any AD & High computational cost \\
\cite{ikeda2018estimation} & S (VAE) & F & Others (True latent distribution based feature contribution) & Static TN & L & More reliable than reconstruction error-based explanations  & Only applicable to VAE \\
\cite{li2021multivariate} & S (VAE) & F & Others (MCMC-based method)  & Streaming MTS & L  & Provides real-time explanations & Unable to handle evolving data\\
\cite{memarzadeh2022multiclass} & A (VAE) & F & Perturbation based feature importance & Static MTS & L & Applicable to any AD & Needs a few labelled data\\
\cite{assaf2021anomaly} & S (ConvAE) & F \& S & Similar historic anomaly + Reconstruction error based feature contribution & Static MTS & L & End-to-end framework & Only applicable to Convolutional AEs \\
\cite{de2021interpretable} & S (AEs) & F & Approximate (Rule extraction from reconstruction error) & Streaming MTS & G  & Handles streaming data & Needs many anomalous samples; Only applicable to reconstruction-based AD \\
\cite{chen2021daemon} & S (AAE) & F & Reconstruction error based feature contribution & Streaming MTS & L & Provides quantitative evaluation of explanation quality & Only applicable to reconstruction-based AD \\
\cite{rajendran2018saife} & S (AAE) & F & Reconstruction errors coupled with the semi-supervised features in AAE & Static MTS & L & Works in both unsupervised and semi-supervised settings & Only applicable to AAE \\
\cite{szymanowicz2022discrete} & S (U-Net based AEs) &  F &  Saliency maps + Human-understandable representation  & Static video & L & Provides visual explanations & Only applicable to AEs \\
\cite{gnoss2022xai} & A (AEs) & F & SHAP + Approximate (Decision tree + Linear regression)  & Static TM & L & Applicable to any AD & Poor explanation fidelity; Manual evaluation of explanation quality\\
\cite{amarasinghe2018toward} & S (DNN) & F & LRP based feature relevance & Any data type & L \& G & Not easy to understand for non-experts & Needs labelled data; Only applicable to certain DNNs \\
\cite{patil2019explainable} & S (LSTM) & F & LRP based feature relevancy & Static ES & L & Not easy to understand for non-experts & Needs labelled data; Only applicable to certain DNNs \\
\cite{tallon2020explainable} & A (DT + LSTM) & F  & Shapley values based feature importance & Static ES &  L & Applicable to any AD & Poor explanation fidelity\\
\cite{nor2021application,hwang2021sfd,jakubowski2021explainable} & A (LSTM) & F & SHAP & MTS & L & Applicable to any AD & Poor explanation fidelity \\
\cite{herskind2021outlier} & A (LSTM) & F & Approximate (LIME) & Streaming UTS & L & Applicable to any AD; Handles streaming data & Not easy to understand for non-experts \\
\cite{han2021interpretablesad} & S (LSTM) & F & Integrated Gradients &  Static ES & L & Not needs labelled data  & Only applicable to DNN; Unable to handle evolving data  \\
\cite{brown2018recurrent} & S (LSTM) & F & Attention mechanism & Streaming ES & L & Handles streaming data &  Only applicable to RNNs\\
\cite{mathonsi2021statistics} & A (MES-LSTM) & F & Approximate (LIME) + SHAP & Static MTS & L & Applicable to any AD & Poor explanation fidelity\\
\cite{saeki2019visual} & S (CNN) & F & grad-CAM + Visualization & Static UTS & L & Provides visual explanations & Only applicable to CNNs\\
\cite{cheong2021interpretable} & A (CNN) & F & Approximate (LIME) & Static UTS & L & Applicable to any AD & Poor explanation fidelity \\
\cite{chong2021toward} & S (CNN) & F \& S & LRP based feature importance + Prototypes explanation + Visualisation & Static image &  L  & Provides visual explanations & Hard to set the number of prototypes\\
\cite{levy2022anomili} & A (AE + CNN + LSTM) & P & Approximate (Decision Tree) + SHAP TreeExplainer & Streaming MTS  & L & Applicable to any AD; Provides real-time detection and explanations & Poor explanation fidelity\\
\cite{szymanowicz2021x} & S (R-CNN) & F & Others (PCA + GMM) + Visualisation & Static video & L & Provides visual explanations & Unable to handle unseen data in training phrase  \\
\cite{hinami2017joint} & S (R-CNN) & F & Others (Semantic Anomaly Score) & Static video & L & Joint abnormal event
detection and recounting & Weak interpretability using only semantic anomaly scores\\
\cite{sipple2020interpretable} &  S (DNN) & F \& S & Integrated Gradients  &  Streaming TS & L & Handles streaming data & Only applicable to DNN \\
\cite{xu2021beyond} & A (DevNet) & F & Others (Self-attention learning based feature selection) &  Static TM & L & Applicable to any AD & Poor explanation fidelity \\
\cite{pang2021explainable} & S (DevNet) & F & Gradient back propagation based feature contribution  & Static image & L & End-to-end training & Only applicable to specific DNN\\
\cite{kauffmann2020towards, kauffmann2020clever} & S (OCSVM) & F & Approximate (NN) + LRP-type Back-propagation based feature importance at pixel-level & Static image & L & Better performance compared to others & Only applicable to OCSVM and potentially some distance-based methods \\
\cite{liu22interpretable} & S (DNN) & F & Attention mechanism + GMM + Visualisation & Static UTS &  L & Handles time series with varying lengths; Provides visual explanations & Only applicable to specific DNN\\
\hline  
\end{longtable}
\end{landscape}

Finally, visualisation techniques can be leveraged to help explain anomalies. For instance, {\color{black}Liu et al. }\cite{liu22interpretable} create the deep temporal clustering framework seq2cluster, which can cluster and detect anomalies in time series with varying lengths. The Temporal Segmentation, Temporal Compression network, and GMM Estimation modules make up seq2cluster. In particular, each sequence is divided into non-overlapping temporal segments via the Temporal Segmentation module. A low-dimensional representation of each time segment is what the Temporal Compression network aims to achieve. Moreover, the Estimation Network for GMMs utilises the latent space representation to perform density estimation. Therefore, data instances can be clustered in latent space to find anomalies based on the likelihood of each segment sample. The results of anomaly detection can also be more easily interpreted when anomalies found in the latent space are adequately visualized. 

\textit{Discussion:} In addition to the above mentioned DNNs, namely AEs, RNNs, CNNs, GANs, Deep OCSVM and DevNet, other DNNs such as Graph Neural Networks \citep{chaudhary2019anomaly} and Transformers \citep{lin2021survey} have become prevalent in anomaly detection. Therefore, the interpretation methods of these DNNs are also relevant.
%However, as of this writing, we did not find any work designing XAD based on these DNNs.

\subsection{Summary}

To wrap up our review of post-model XAD techniques for deep neural networks, Table~\ref{tab:summary_deep_post} gives an overview of all techniques discussed and we have several high-level observations.

First, most deep post-model XAD techniques are model-specific in the sense that they are only applicable to a family of specific neural networks or all neural networks. This is in stark contrast with most shallow post-model XAD techniques, which are typically model-agnostic. This is because these deep post-model XAD techniques provide explanations by exploring the internal structure of the neural network. By doing so, although these explanation methods cannot be generalized to other anomaly detection models, the resulting explanations are usually faithful as the \textit{Explanation-Definition} is in compliance with the \textit{Detection-Definition}. However, techniques such as SHAP, LIME, and some rule learners are model-agnostic and are therefore more likely to suffer from poor fidelity.

Second, nearly all deep post-model XAD techniques provide only feature-based explanations; the only exceptions are References \citep{giurgiu2019additive,assaf2021anomaly,chong2021toward}, which also produce sample-based explanations. Regarding the techniques used, the Shapley values based approach is the most popular one. More importantly, one can see that most deep post-model XAD techniques are proposed to explain anomalies detected in sequential data such as time series and system logs.

Third, nearly all deep post-model XAD techniques can only provide local explanations. In other words, they can only explain a single anomaly at a time. Due to the complexity of neural networks, it is extremely challenging, if possible, to understand the entire decision-making process. To help end-users understand why an instance is reported as anomalous, deep post-model XAD techniques often inspect some important properties of the neural networks, such as feature contribution to  reconstruction errors, thereby providing weak interpretability.

\section{Conclusion and Future Opportunities}

We reviewed more than 150 papers that harness XAD techniques to explain anomalies. Specifically, we first introduced three different definitions of anomaly, and then clarified what XAD is and why it is needed. On this basis, and inspired by existing surveys on XAI, we proposed a taxonomy consisting of six main criteria, enabling the categorization of the increasingly rich field of XAD. For purposes of brevity and organisation, we structured the review into four high-level categories (corresponding to sections S4--7) and twelve fine-grained categories (corresponding to the subsections of S4--7). Throughout the survey we identified a number of research challenges that may offer opportunities for future research, which we will summarize next.

\subsection{Definition of Anomaly and XAD}
A long-standing problem in anomaly analysis is the lack of a uniform definition of an anomaly, leading to a wide range of anomaly detection methods. The diversity of anomaly definitions and anomaly detection methods leads to the need for a large variety of anomaly explanation methods. Although is not necessarily problematic on itself (and may be unavoidable), the lack of uniform definitions for anomaly detection and XAD hampers communication of researchers between different (sub)fields, such as computer vision, natural language processing, data mining, and social science. This makes it hard to find related work and leads to the re-invention of methods, causing unnecessary delays in scientific progress. More importantly, the evaluation and comparison of XAD methods becomes difficult and subjective, due to the lack of a uniform, objective, and precise definition of XAD.

\subsection{Evaluation of XAD }

Despite the clearly stated needs for XAD in various domains, and especially those domains involving high-stakes decisions, the question of how XAD techniques should be evaluated remains unanswered. Over the past few years, the long-standing problem of measuring and assessing machine learning explainability has received certain attention \citep{carvalho2019machine}. However, most of these methods are specifically designed for classification or clustering problems, and extending these methods to anomaly detection problems is non-trivial.

%\subsection{Fidelity of Post-hoc Explanations}

Particularly, the fidelity of post-hoc explanations merits close scrutiny when evaluating XAD techniques. Inconsistency between the \textit{Oracle-Definition} and \textit{Detection-Definition} of an anomaly may lead to the identification of anomalies that are not of interest to the end-users. Moreover, inconsistency between the \textit{Detection-Definition} (if available) and \textit{Explanation-Definition} of an anomaly may lead to poor explanation fidelity. In other words, the explanations do not reflect the actual decision-making process of an anomaly detection model. In general, pre-model and in-model XAD techniques do not suffer from this problem, while most post-hoc XAD techniques---that only correlate inputs with outputs, without exploring the internal structure of detection models---are afflicted with this problem. For this reason, some researchers \citep{rudin2019stop} appeal not to use opaque models and then explain them in a post-hoc manner for high-stakes decisions. Instead, an intrinsically explainable model should be used. We emphasize that the same argument also applies to anomaly detection and explanation.

\subsection{XAD with Prior Knowledge}
Most XAD techniques attempt to provide explanations solely based on information contained in existing data instances (anomalous or not) and/or anomaly detection models. However, sometimes additional knowledge about data instances or anomaly detection models may be acquired, in the form of algebraic equations, simulation results, logic rules, knowledge graphs, human feedback, etc. Importantly, this prior knowledge can be integrated to augment training data, choose a network architecture, initialize parameters or validate model outputs. This paradigm of integrating prior knowledge into  machine learning is called \textit{Informed Machine Learning} \citep{von2019informed}, which has received increasing attention over the past few years. Particularly, {\color{black}Beckh et al. }\citep{beckh2021explainable} have performed a survey on methods that integrate prior knowledge into machine learning for improving explainability, wherein they subdivide these methods into three categories, including the integration of knowledge into machine learning problems such as classification, regression, clustering or anomaly detection, the integration of knowledge into explanation method, and deriving knowledge from the explanation results and then integrating it into the machine learning pipeline. Therefore, repurposing these methods for anomaly detection is undoubtedly beneficial to improve interpretability, and thus is a promising future direction.

\subsection{Adversarial Attacks in XAD}
{\color{black}Belle \& Papantonis }\cite{belle2021principles} point out that some  widely used XAI techniques are vulnerable to adversarial attacks. In particular, post-hoc XAI techniques such as LIME and SHAP are easily manipulated \citep{slack2020fooling}. From the reviewed results, we can see that most of these methods in XAI have been repurposed for XAD. As a result, anomaly explanations obtained by using these techniques have the possibility of being manipulated or attacked. To circumvent this problem, attention should be paid when selecting an XAD method. In particular, more efforts should be made to develop interpretation methods that take into account adversarial attacks in the future.

\subsection{Scalability of XAD}
Last but not least, the scalability of XAD plays an important role in some applications. For example, large internet companies often develop an anomaly detection system to monitor a large number of key performance indicators, aiming to ensure the reliability of their service platform \citep{li2018robust}. However, after identifying anomalies, they have to find the root causes and then take remedial actions as soon as possible. To achieve this in an automated manner, an anomaly interpretation method that can process large amounts of data and provide near real-time interpretation is required. However, most existing XAD techniques---such as subspace anomaly detection and Shapley value based methods---have a high computational cost. Therefore, the development of scalable XAD techniques (with high fidelity) is an important direction for future research.

\begin{acks}
This publication is part of the project Digital Twin with project number P18-03 of the research programme TTW Perspective, which is (partly) financed by the Dutch Research Council (NWO). We thank Dr. Gabriel de Albuquerque Gleizer for his valuable feedback.
\end{acks}

\bibliographystyle{ACM-Reference-Format}
\bibliography{sample-base}

%\appendix

\end{document}